\pdfoutput=1

\documentclass[11pt]{article}

\usepackage[]{emnlp2021}

\usepackage{times}
\usepackage{latexsym}
\usepackage{graphicx}
\usepackage{caption}
\usepackage{subcaption}
\usepackage{soul}
\usepackage{todonotes}

\usepackage[T1]{fontenc}

\usepackage[utf8]{inputenc}

\usepackage{microtype}
\usepackage{lipsum}
\usepackage{adjustbox}

\newcommand{\systemname}{\textsc{arae}$_{\textit{seq2seq}}$}  
\newcommand{\drg}{\textsc{drg}}  
\newcommand{\arae}{\textsc{arae}} 
\newcommand{\yelpdata}{\textsc{yelp}}  
\newcommand{\imdbdata}{\textsc{imdb}}  
\newcommand{\politicaldata}{\textsc{political}} 
\newcommand{\srcdom}{$\mathcal{S}$}
\newcommand{\trgdom}{$\mathcal{T}$}
\definecolor{darkgreen}{HTML}{5EE261}
\definecolor{darkorange}{HTML}{F8CAAD}
\definecolor{red}{cmyk}{0, 1, 1, 0}
\definecolor{algohighlight}{HTML}{000000}
\definecolor{Gray1}{rgb}{0.91,0.925, 0.937}
\definecolor{Gray2}{rgb}{0.87, 0.886, 0.902}
\definecolor{Gray3}{rgb}{0.808, 0.831, 0.855}
\definecolor{Gray4}{rgb}{0.678,0.71, 0.741}
\definecolor{Green0}{rgb}{0.909, 0.992, 0.886}
\definecolor{Green1}{rgb}{0.843, 0.960, 0.839}
\definecolor{Green2}{rgb}{0.635, 0.854, 0.627}
\definecolor{Red1}{rgb}{0.937, 0.686, 0.698}

\definecolor{darkgreen}{rgb}{0.0, 0.5, 0.0}
\definecolor{almond}{rgb}{0.99, 0.87, 0.9}
\definecolor{ghostwhite}{rgb}{0.97, 0.97, 1.0}
\definecolor{Blue1}{rgb}{0.792, 0.941, 0.973}
\definecolor{Blue2}{rgb}{0.678, 0.91, 0.957}
\definecolor{Blue3}{rgb}{0.565, 0.878, 0.937}
\definecolor{Blue4}{rgb}{0.282, 0.749, 0.89}

\definecolor{Yellow1}{rgb}{1, 0.914, 0.306}
\definecolor{Yellow2}{rgb}{1, 0.886, 0.275}
\definecolor{Yellow3}{rgb}{1, 0.855, 0.239}

\newcommand{\flmetric}{\textsc{fl}}
\newcommand{\simmetric}{\textsc{sim}}
\newcommand{\accmetric}{\textsc{acc}}
\newcommand{\aggmetric}{\textsc{agg}}
\newcommand{\abhi}[1]{\textcolor{black}{#1}}

\usepackage[]{amsmath}
\usepackage{booktabs}

\usepackage{multirow}
\usepackage{multicol}
\usepackage{graphicx}
\usepackage{makecell}
\usepackage{amsfonts}
\usepackage{color,soul}
\usepackage{enumitem}

\usepackage[ruled,vlined,linesnumbered,scleft]{algorithm2e}
\SetCommentSty{mycommfont}
\DontPrintSemicolon
\usepackage{xcolor} 
\usepackage{cellspace}
\usepackage{url}
\usepackage[compact]{titlesec}
\usepackage[hang,flushmargin]{footmisc}
\usepackage{cleveref} 
\crefname{section}{§}{§§}
\Crefname{section}{§}{§§}

\newcommand{\declarelogo}[0]{\includegraphics[height=.02\textwidth]{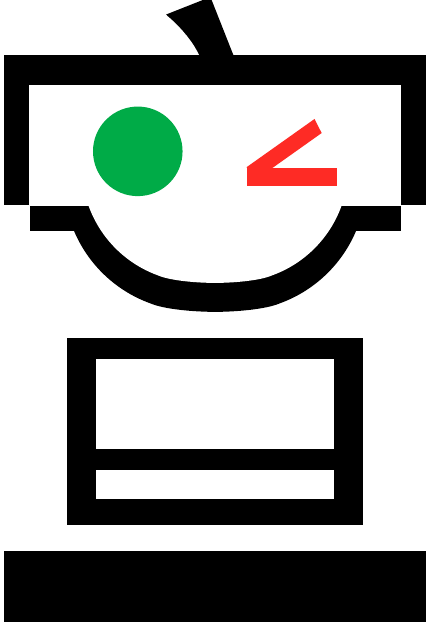}}

\title{So Different Yet So Alike! \\  Constrained Unsupervised Text Style Transfer}

\author{Abhinav Ramesh Kashyap\(^{\alpha*}\), Devamanyu Hazarika\(^{\alpha}\)\thanks{\ \ The first two authors contributed equally} ,\\ \textbf{Min-Yen Kan\(^\alpha\), Roger Zimmermann\(^\alpha\), Soujanya Poria}$^{\declarelogo}$\\
\(^\alpha\) National University of Singapore, Singapore\\
$^{\declarelogo}$ DeCLaRe Lab, Singapore University of Technology and Design, Singapore \\
\fontsize{10}{10}\texttt{\{abhinav, hazarika, kanmy, rogerz\}@comp.nus.edu.sg}\\ 
\fontsize{10}{10}\texttt{sporia@sutd.edu.sg}\\ 
}

\begin{document}
\maketitle
\begin{abstract}
Automatic transfer of text between domains has become popular in recent times. One of its aims is to preserve the semantic content of text being translated from source to target domain. However, it does not explicitly maintain other attributes between the source and translated text, for e.g., text length and descriptiveness. Maintaining constraints in transfer has several downstream applications, including data augmentation and de-biasing.  
We introduce a method for such constrained unsupervised text style transfer by introducing two complementary losses to the generative adversarial network (GAN) family of models. Unlike the competing losses used in GANs, we introduce cooperative losses where the discriminator and the generator cooperate and reduce the same loss. The first is a \textit{contrastive} loss and the second is a \textit{classification} loss --- aiming to regularize the latent space further and bring similar sentences across domains closer together. We demonstrate that such training retains lexical, syntactic, and domain-specific constraints between domains for multiple benchmark datasets, including ones where more than one attribute change.  We show that the complementary cooperative losses improve text quality, according to both automated and human evaluation measures. \footnote{\url{https://github.com/abhinavkashyap/dct}}
\end{abstract}

\section{Introduction}



Modern neural networks methods are capable of mapping data from one domain to another. Prominent examples include translation of text between languages~\cite{DBLP:conf/nips/VaswaniSPUJGKP17,artetxe2018unsupervised,lample_unsupervised_mt}, emoji creation from human faces \cite{DBLP:conf/iclr/TaigmanPW17}, and stylistic transfer of speech~\cite{yuan2021improving}. In Natural Language Processing (NLP),  the umbrella term \textit{attribute transfer}~\cite{jin2020deep} (or \textit{domain transfer}) refers to similar methods\footnote{While the literature primary utilizes the term \textit{style transfer}, we adopt the more general term \textit{attribute} as suggested by \citet{dijin-survey}.}. The aim is to maximally preserve the semantics of the source sentence (``content'') but change other properties (``attributes''), such as sentiment \cite{jin2020deep}, expertise \cite{cao-etal-2020-expertise}, formality  \cite{formality-style-rao} or a combination of them \cite{multiple-attribute-text-style-transfer}.

\begin{figure}
    \centering
    \includegraphics[width=0.9\columnwidth]{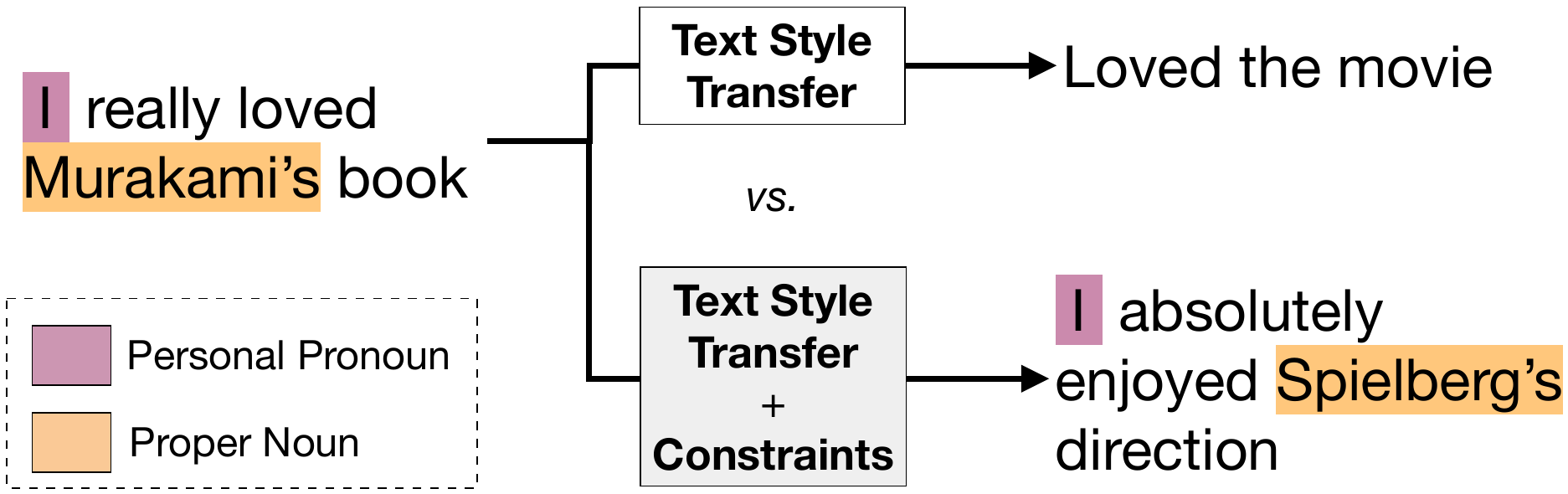}
    \caption{\footnotesize Illustrative example showing transfer of text from books to movies while maintaining constraints of identity. 
    }
    \label{fig:illustrative_example}
\end{figure}

\textit{Text style transfer}, a popular form of attribute transfer, regards ``style'' as any attribute that changes between datasets \cite{dijin-survey}. Building on the progress of supervised transfer models, recent works have focused on \textit{unsupervised style transfer} that avoids costly annotation of parallel sentences. 
However, models built using unsupervised methods perform poorly when compared to supervised (parallel) training \cite{artetxe-etal-2020-call}. These methods, while capable of achieving the target domain characteristics, often fail to maintain the invariant content. \Cref{fig:illustrative_example} illustrates one such example, where a sentence from the 
{\sc Books} domain is translated to the {\sc Movie} domain. While the translated sentence ``\textit{Loved the movie}'' has correctly transferred the attribute (style), it does not have the same length, does not retain the personal noun (``\textit{I}''), nor use a domain-appropriate proper noun. Comparatively, the higher-fidelity transfer ``\textit{I absolutely enjoyed Spielberg's direction}'', maintains such \textit{constraints of identity}, in addition to being apt.


This problem setting is an important application of text transfer, as 
enforcing constraints of identity can help maintain the brand identity when the product descriptions are mapped from one commercial product to another. They can also help in data augmentation for downstream domain adaptation NLP applications (\cref{section:discussion}). Constraints of identity are explored extensively in the computer vision task of cross-domain image generation.
\cite{DBLP:conf/iclr/TaigmanPW17}, but these issues---to the best of our knowledge---are  unexplored in NLP.

In this paper, we improve unsupervised attribute transfer by enforcing invariances via explicit constraints. 
Current methods in text attribute transfer lack mechanisms to explicitly enforce such constraints between the source and the transferred sentence. To this end, we build upon unsupervised text style transfer work by introducing an additional explicit regularization component in the latent space of a GAN-based \textit{seq2seq} network through two complementary losses (\cref{sec:method}). Unlike the adversarial losses in the GAN framework, our proposed losses cooperatively reduce the same objective. The first loss is a contrastive loss \cite{contrastive_learning_review} that brings sentences that have similar constraints closer and pushes sentences that are dissimilar farther away. The second loss is a classification loss that helps maintain the sentence identity via constraints from the latent vectors \cite{acgan}.

Our approach, while simple and aimed at maintaining constraints, improves the overall performance of the generation. We demonstrate these gains over three datasets: \yelpdata{} \cite{DBLP:conf/icml/ZhaoKZRL18}, \imdbdata{} \cite{dai-etal-2019-style} and \politicaldata{} \cite{prabhumoye-etal-2018-style}, generating six constraints including lexical, syntactic and domain-specific constraints. The introduced cooperative losses satisfy the constraints more effectively compared against strong baselines. Since multiple attributes can change between two domains \cite{multiple-attribute-text-style-transfer}, we test our method on one such dataset and show that the constraints of identity are maintained more effectively (\Cref{sec:results-maintaining-constraints}). 
To the best of our knowledge, our approach is the first to introduce cooperative losses in a GAN-like setup for NLG. 





\section{Preliminaries}

\paragraph{Task Setup:} We consider two sets of sentences (or corpora) \srcdom{}$=\{x_{src}^1, x_{src}^2, \dots x_{src}^{m}\}$ and \trgdom{}$=\{x_{trg}^1, x_{trg}^2, \dots x_{trg}^{n}\}$, as the \textit{source} and \textit{target} domains, respectively. Each corpus --- which we interpret as domains --- contain discernable attributes, ranging from sentiment (e.g., positive vs. negative), topics, political slant (e.g., democratic vs. republican), or some combination \cite{li-etal-2018-delete,DBLP:conf/iclr/LampleSSDRB19}. The overall task is to rewrite a piece of text $s_i \in \text{\srcdom{}}$ to $t_i \in \text{\trgdom{}}$, such that the translation changes the attributes varying across the two domains but retains the remaining content. While content retention is not explicitly defined in the literature, we design this new task of constrained unsupervised attribute transfer that assigns explicit constraints $\mathcal{C} = \{c_1, c_2, \ldots, c_{|\mathcal{C}|}\}$, to be retained. These constraints can be defined at various levels of a sentence: lexical, syntactic and domain-specific.

\begin{figure*}[t!]
    \centering
    \subfloat[ \label{fig:vanilla-dct}]{\includegraphics[width=0.45\textwidth]{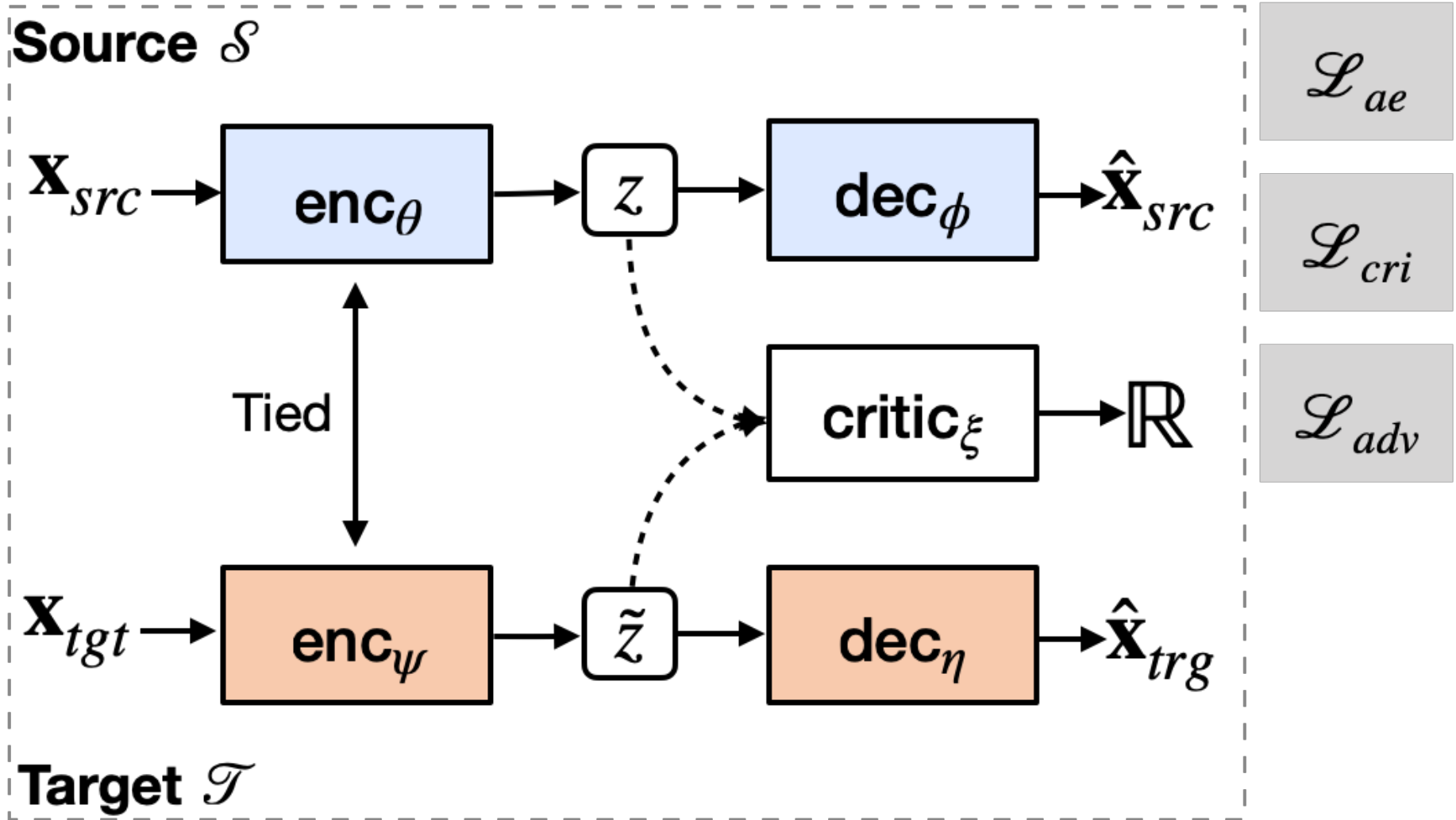}} 
    \hspace{0.8cm}
    \subfloat[\label{fig:contra-clf-dct}]{\includegraphics[width=0.45\textwidth]{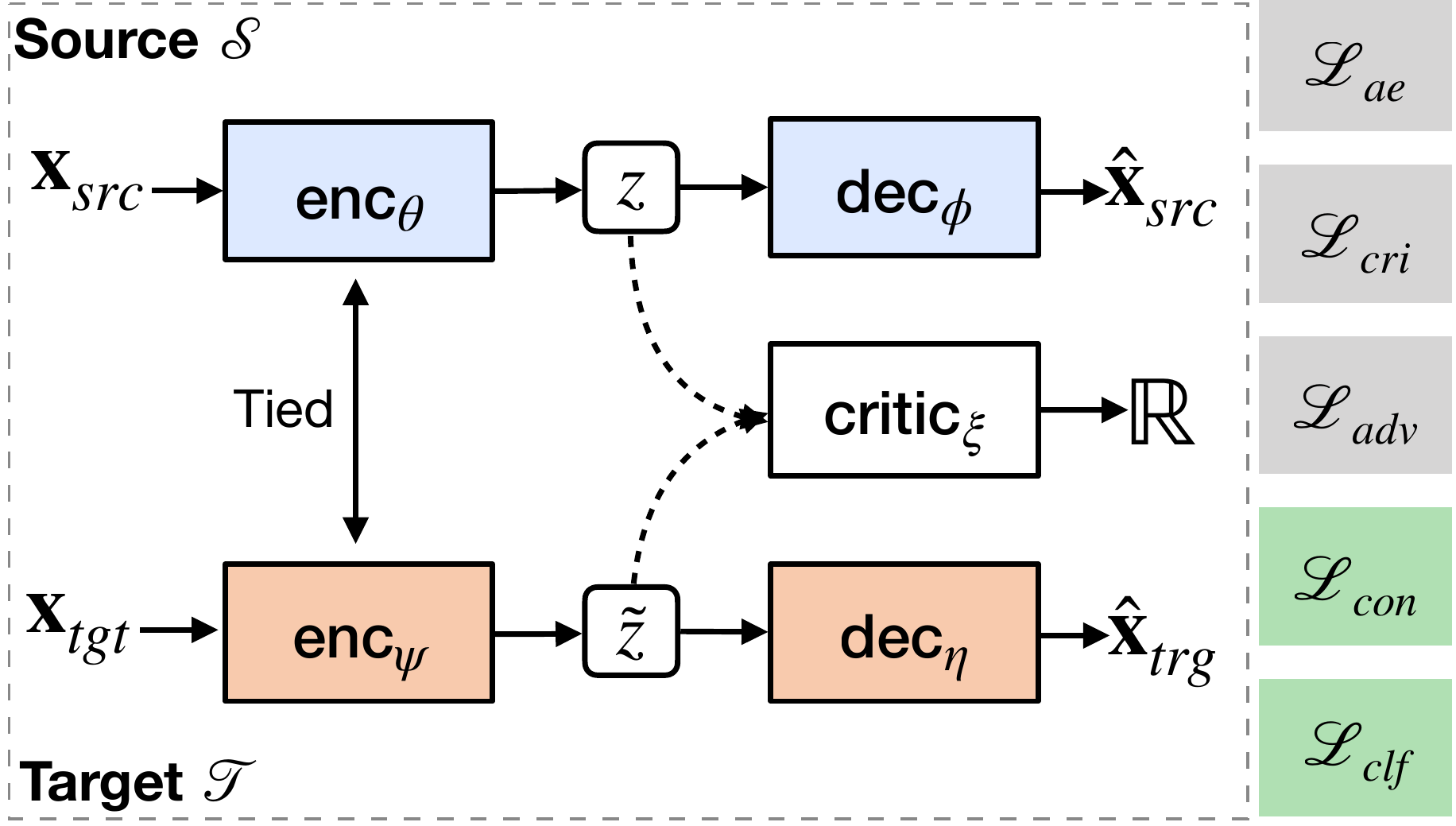}} 
    \caption{(a) \systemname{} -- We replace the generator of \arae{} with an encoder that encodes text from $\mathcal{T}$. (b) Adding our proposed cooperative losses to the model.}
    \label{fig:architecture}
\end{figure*}

\paragraph{Adversarially Regularized Autoencoder} (\arae{}): To perform unsupervised attribute transfer, we consider \textit{seq2seq} models that encode source sentences to a latent space and then decodes them to the target sentences. \arae{}s \cite{DBLP:conf/icml/ZhaoKZRL18} are the auto-encoder variants of the Generative Adversarial Network (GAN) \cite{DBLP:journals/corr/GoodfellowPMXWOCB14} framework. They learn smooth latent spaces (by imposing implicit priors) to ease the sampling of latent sentences. \arae{}s have been widely adopted in tasks like unsupervised text generation \cite{huang-etal-2020-cycle}, topic modeling \cite{hu-etal-2020-neural}, among others, and form the backbone of our proposed model.

\arae{} consists of an auto-encoder with a deterministic encoder $\displaystyle enc_{\theta}: \mathcal{X} \rightarrow \mathcal{Z}$ that encodes sentences into a latent space; i.e., $\mathbf{z} = enc_{\theta}(\mathbf{x}) \sim P_{z}$, and a conditional decoder $p_\phi(\mathbf{x}|\mathbf{z})$ that generates a sentence given a latent code. \arae{} regularizes this latent space utilizing a GAN-like setup that includes an implicit prior obtained from a parameterized generator network $enc_{\psi}: \mathcal{N}(0,I) \rightarrow \mathcal{Z}$. Here,  $enc_{\psi}$ maps a noise sample $s \sim \mathcal{N}(0,I)$ to the corresponding prior latent code $\mathbf{\bar{z}} = enc_{\psi}(s) \sim P_{\bar{z}}$.

A critic $\displaystyle crc_{\xi}: \mathcal{Z}  \rightarrow \mathbb{R}$ then learns to distinguish between real and generated samples, whereas both $enc_{\theta}$ and  $enc_{\psi}$ are adversarially trained to fool the critic. This results in a minimax optimization which implicitly minimizes the JS-Divergence between the two distributions $\displaystyle P_{z}$ and $\displaystyle P_{\bar{z}}$:

\begin{equation}
    \underset{\psi}{\min} \ \underset{\xi}{\max}  \quad 
    \underset{\mathbf{z} \sim P_{z}}{\mathbb{E}}[crc_{\xi}(\mathbf{z})] - \underset{\mathbf{\bar{z}} \sim P_{\bar{z}}}{\mathbb{E}}[crc_{\xi}(\mathbf{\bar{z}})] \label{Eq:original-minimax}
\end{equation}

The training involves three optimizations: $i)$ reducing the auto-encoder loss $\mathcal{L}_{ae}$, which tries to reconstruct the input and encourages copying behavior and maintain semantics similar to the original text (Eq.~\ref{Eq:autoencoder-training}); $ii)$ optimizing the critic's loss $\mathcal{L}_{cri}$ to distinguish between real and fake samples (Eq.~\ref{Eq:disc-training}) $iii)$ training the encoder and generator $\mathcal{L}_{adv}$ to fool the critic (Eq.~\ref{Eq:adv-training}):

\begin{eqnarray}
    \mathcal{L}_{ae}(\theta, \phi)=\underset{\mathbf{z} \sim P_{z}}{\mathbb{E}}[-\log\ p_{\phi}(\mathbf{x} | \mathbf{z})]  \label{Eq:autoencoder-training} \\
    \mathcal{L}_{crc}(\xi)=\underset{\mathbf{z} \sim P_{z}}{-\mathbb{E}}[ crc_{\xi}(\mathbf{z}) ] + \underset{\mathbf{\bar{z}} \sim P_{\bar{z}}}{\mathbb{E}} [crc_{\xi}(\mathbf{\bar{z}}) ]  \label{Eq:disc-training} \\
    \mathcal{L}_{adv}(\theta, \psi)=\underset{\mathbf{z} \sim P_{z}}{\mathbb{E}}[crc_{\xi}(\mathbf{z}) ] - \underset{\mathbf{\bar{z}} \sim P_{\bar{z}}}{\mathbb{E}}[crc_{\xi}(\mathbf{\bar{z}}) ] \label{Eq:adv-training} 
\end{eqnarray}

\section{Proposed Method} \label{sec:method}

\subsection{Base Model (\systemname{})} While \arae{} is an auto-encoder that recreates input $\mathbf{x} \rightarrow \mathbf{\hat{x}}$, our requirement is to translate sentences from one domain to another. Given this, we modify the \arae{} to a \textit{seq2seq} variant such that we can translate input sentences between source and target domains; i.e., $\mathbf{x}_{src} \rightarrow \mathbf{\hat{x}}_{tgt}$ and $\mathbf{x}_{tgt} \rightarrow \mathbf{\hat{x}}_{src}$.

To achieve this, we  utilize $enc_\theta$ to encode $\mathbf{x}_{src}$ and repurpose $enc_\psi$ to encode $\mathbf{x}_{tgt}$. We obtain their latent codes $(\mathbf{z},\mathbf{\bar{z}})$ which we name as $(\mathbf{z}^s,\mathbf{z}^t)$, i.e., $\mathbf{z}^s = enc_\theta(\mathbf{x}_{src})$ and $\mathbf{z}^t = enc_\psi(\mathbf{x}_{tgt})$.

Next, to generate sentences, we consider two decoders $\mathbf{\hat{x}}_{src} \sim p_\phi(\mathbf{x}|\mathbf{z})$ and $\mathbf{\hat{x}}_{tgt} \sim p_\eta(\mathbf{x}|\mathbf{z})$. Here, $\mathbf{z}$ can be either $\mathbf{z}^s$ or $\mathbf{z}^t$ based on whether we auto-encode (e.g., $p_\phi\left(\mathbf{x}|\mathbf{z}^s=enc_\theta(\mathbf{x}_{src})\right)$) or translate (e.g., $p_\phi\left(\mathbf{x}|\mathbf{z}^t=enc_\psi(\mathbf{x}_{tgt})\right)$). Unlike \arae{}'s single decoder, we incorporate two decoders to enable bi-directional translation.

In the above process, instead of sampling $s$ from a noise distribution like $\mathcal{N}(0, I)$ and passing it through a generator $enc_{\psi}$, we feed it text from the target domain \trgdom{} and a decoder $dec_{\eta}$ that decodes text in \trgdom{}. This is inspired from Cycle-GAN \cite{DBLP:conf/iccv/ZhuPIE17}, where instead of matching the noise distribution $\mathcal{N}$, we match the distribution of \trgdom{}.

In addition, we tie the weights of the encoders from both domains, so that the encoders learn to encode domain-agnostic information. Tying encoder weights has also been used by unsupervised machine translation \cite{artetxe2018unsupervised,lample_unsupervised_mt}
and multiple other works \cite{DBLP:conf/emnlp/MaiPMSH20, huang-etal-2020-cycle,hu-etal-2020-neural, artetxe2018unsupervised}\footnote{We tried with separate encoders and decoders, but encoders with tied weights work best}. 

\begin{algorithm}[t!]

 \For{each training iteration} {
    \colorbox{Gray2}{1) Train the Auto-encoders:} \\ 
       {\Indp Sample $\; \mathbf{x}_{src} \sim \mathcal{S} \;$,
        $\; \mathbf{x}_{trg} \sim \mathcal{T}$ \\ 
        $\mathbf{z}^s = enc_{\theta}(\mathbf{x}_{src}),\ \mathbf{z}^t = enc_{\psi}(\mathbf{x}_{trg})$ \\ 
        Backprop loss, $\mathcal{L}_{ae}(\theta, \phi)$, $\mathcal{L}_{ae}(\psi, \eta)$\\}
        
    \colorbox{Green2}{2) Train the Critic:} \\
        
        {\Indp Sample $\; \mathbf{x}_{src} \sim \mathcal{S} \;$,
        $\; \mathbf{x}_{trg} \sim \mathcal{T}$ \\ 
        $\mathbf{z}^s = enc_{\theta}(\mathbf{x}_{src}),\ \mathbf{z}^t = enc_{\psi}(\mathbf{x}_{trg})$ \\
        $\mathbf{z}_{crc}^s = crc_{\xi}^{hid}(\mathbf{z}^s),~\mathbf{z}_{crc}^t = crc_{\xi}^{hid}(\mathbf{z}^t)$ \\ 
    
        $l_{crc} \leftarrow \mathcal{L}_{crc}(\xi)$ \\ }
        
    \colorbox{Green1}{2a) Critic Co-op Training:} \\ 
        
        {\Indp 
        Backprop loss, $l_{crc} + \lambda_1\mathcal{L}_{con}(\xi) + \lambda_2\mathcal{L}_{clf}(\xi,~\delta)$ \\
        }
    
    \colorbox{Blue2}{3) Adversarial Training:} \\
        {\Indp 
        Sample $\; \mathbf{x}_{src} \sim \mathcal{S} \;$,
        $\; \mathbf{x}_{trg} \sim \mathcal{T}$ \\ 
        $\mathbf{z}^s = enc_{\theta}(\mathbf{x}_{src}),\ \mathbf{z}^t = enc_{\psi}(\mathbf{x}_{trg})$ \\ 
        Backprop loss, $\mathcal{L}_{adv}(\theta, \psi)$\\ 
        }
        
    \colorbox{Blue1}{3a) Encoder Co-op Training:} \\ 
        {\Indp 
        Backprop loss, $\lambda_1\mathcal{L}_{con}(\theta,~\phi) + \lambda_2\mathcal{L}_{clf}(\theta,~\phi,~\delta)$ \\
        }
 }
\caption{\systemname{} + \textsc{clf} + \textsc{contra}}
\label{algo:training-algo}
\end{algorithm}

\subsection{Adding Constraints via Co-op Training}

While the latent space in \systemname{} learns to match \srcdom{} and \trgdom{} sentences, there is no guarantee on translations maintaining the 
``content''. This issue is particularly pronounced in unsupervised attribute transfer due to lack of parallel sentences between \srcdom{} and \trgdom{}.

To alleviate the issue, we propose to learn a structured latent space which embodies notions of our constraints in its embedded latent codes.  This ensure that instances with similar constraints are closer in the latent space. In particular, we propose two types of optimization --- self-supervised and discriminative --- to maintain the constraints better.

\subsubsection{Cooperative Contrastive Learning}

We use contrastive representation learning to regularize the latent space, such that encoders bring two sentences sharing similar constraints closer together (positive pairs), and force dissimilar ones away (negative pairs). For example, sentences of similar lengths (irrespective of their domains) should be closer together.

Among many self-supervised metric losses such as Triplet Loss \cite{tripletloss} and NT-Xent loss \cite{DBLP:conf/icml/ChenK0H20}, we use one that is amenable to multiple positive instances \cite{DBLP:conf/nips/KhoslaTWSTIMLK20}. Given a sentence $s_i \in \mathcal{S}$ in a mini-batch of size $B$, we mine $P$ positive sentences each from \srcdom{} and \trgdom{} that share the same constraints with $s_i$. This contrastive loss is given by:

\small 
\begin{equation}
    \mathcal{L}_{con}(\theta, \psi, \xi) = -\frac{1}{|P|}\log \left( 
        \sum_{j=1}^{P} 
        \frac{
        e^{(\mathbf{z}_i \cdot \mathbf{z}_j)}}{
        \sum_{k=1}^{B\setminus \{i\}} e^{(\mathbf{z}_i \cdot \mathbf{z}_k)}
        }
    \right)
    \label{eq:contra-loss}
\end{equation}

\normalsize 
\noindent where $\mathbf{z}$'s are representations obtained from the encoders in \srcdom{}, \trgdom{}  or representations obtained from the last layer of critic $crc_{\xi}$.  $\mathcal{C}_i$ are a set of constraints for a sentence. Recently, \cite{DBLP:conf/nips/KangP20} introduced the cooperative loss in the adversarial setup where contrastive losses are added to both the \textit{critic} and \textit{generator} for GANs. Unlike the normal opposing losses of the generator and the critic, both of them cooperatively reduce the contrastive loss. We follow a similar principle and add the loss to both the encoders and the critic (Lines~18).

\subsubsection{Cooperative Classification}
Contrastive learning might be sub-optimal if we do not mine good quality positive and negative samples \cite{DBLP:conf/nips/Tian0PKSI20}. To address this, we propose another way to regularize the latent space. Similar to ACGAN~\cite{acgan}, we encourage the encoders and the critic to cooperatively reduce a classification loss. 
We include a classifier $D_{\delta}:\mathcal{Z}\rightarrow\mathbb{R}^{|\mathcal{C}|}$ that predicts the different constraints $\mathcal{C}$ of the sentences
and the binary cross entropy loss is reduced.


\small 
\begin{align}
\mathcal{L}_{clf}({\theta, \phi, \xi, \delta}) = -\sum_{c=1}^{|\mathcal{C}|} \log{\left(\sigma\left(l_c\right)^{y_c}\left(1 - \sigma\left(l_c\right)\right)^{1-y_c}\right)}
\end{align}

\normalsize
where $|\mathcal{C}|$ is the number of constraints per sentence, $\sigma$ is the sigmoid function and $l_c$ are the logits produced by the classifier for $z_i$. As in contrastive loss, the $z_i$ can be produced by encoders of \srcdom{}, \trgdom{} or from the hidden layers of the critic.

The overall training process is highlighted in~\Cref{algo:training-algo} where $\mathcal{L}_{con}$ and $\mathcal{L}_{clf}$ are weighted by $\lambda_1$ and $\lambda_2$. We choose $\lambda_1,~\lambda_2 \in \{0, 1\}$.


\section{Experiments}
\begin{table}[t!]
    \centering
    \resizebox{\columnwidth}{!}{
    \begin{tabular}{|l l c c c c c|}
        \hline
        \textbf{Dataset} & \textbf{Attributes} & \textbf{Train} & \textbf{Dev} & \textbf{Test} & \makecell{\textbf{Avg} \\\textbf{len.}} & \textbf{Vocab}   \\
        \hline
        \multirow{2}{4em}{\yelpdata}  & Positive & 266,041 & 25,278 & 50,278 & \multirow{2}{1em}{8.9}  & \multirow{2}{1em}{10K} \\
         & Negative & 177,218 & 38,205 & 76,392 &  &\\
        \hline
        \multirow{2}{4em}{\imdbdata}  & Positive & 178,869 & 2K & 1K & 
         \multirow{2}{1em}{18.5} & \multirow{2}{1em}{30K} \\
         & Negative & 187,597 & 2K & 1K & & \\
        \hline
        \multirow{2}{4em}{\politicaldata}  & Democratic & 270,000 & 2K & 28K & \multirow{2}{1em}{16} & \multirow{2}{1em}{30K} \\
         & Republican  & 270,000 & 2K & 28K & & \\
        \hline
    \end{tabular}
    }
    \caption{Dataset splits for \yelpdata, \imdbdata~and \politicaldata. }
    \label{tab:dataset-splits}
\end{table}

\begin{table*}[t!]
    \centering
    \resizebox{0.99\linewidth}{!}{
    \begin{tabular}{|l|l| c c c c | c c c c| c c c c|}
         \hline
          && \multicolumn{4}{c}{\textbf{\yelpdata{}}} & \multicolumn{4}{c}{\textbf{\imdbdata{}}} & \multicolumn{4}{c|}{\textbf{\politicaldata{}}} \\
         \textbf{Model}&\textbf{Sampling} & \textbf{\textsc{acc}} & \textbf{\textsc{fl}} & \textbf{\textsc{sim}} & \textbf{\aggmetric{}} & \textbf{\textsc{acc}} & \textbf{\textsc{fl}} & \textbf{\textsc{sim}} & \textbf{\aggmetric{}} & \textbf{\textsc{acc}} & \textbf{\textsc{fl}} & \textbf{\textsc{sim}} & \aggmetric{} \\
         
         \toprule[0.5pt]  
        
         \hline
         \textbf{\textsc{drg}}&greedy & 67.4 & 54.5 & \textbf{43.6} & 16.7 & 56.5 & 44.3 & \textbf{54.1} & 14.4 & 61.3 & 35.7 & 38.7 & 8.8 \\ 
         
        \textbf{\textsc{arae}}& greedy & \textbf{93.1} & 67.9 & 31.2 & 19.8 & 95.0 & 76.3 & 26.4 & 19.9 & 63.0 & 72.1 & 17.3 & 11.0 \\
         
         \hline
         \hline
         \multirow{2}{*}{\bf \systemname{}} & greedy  & 88.3 & 66.0 & 34.4 & 20.2 & 95.4 &	70.5 &	\textbf{36.4} &	26.0  & 95.80	& 53.1	& 28.5 & 	14.1  \\ 
         
         & nucleus($p=0.6$) & 86.7 & 63.9 & 35.3 & 19.9 &  95.1 &	69.8 &	\textbf{36.4} &	25.6  & 95.8	& 52.2 & 	28.4 & 	13.9   \\ 
         
         
         \hline
         \hline

         \multirow{2}{*}{\makecell[l]{\bf \systemname{} \\ \hspace{0.5cm} +  \bf \textsc{clf}}} &greedy & 85.7	 & 63.4	 & 36.7 & 20.2 &  96.0 & 73.6	& 35.4	& 26.2  & 98.6	& 55.0	& \textbf{44.4}	& \textbf{25.5} \\ 
         
         & nucleus($p=0.6$) & 85.6 & 63.0 & 36.6 & 20.0 & 95.8	& 72.8	& 35.3	& 25.7  & 98.6 &	54.4 &	44.2 &	25.1 \\ 
         

         \hline
         \hline
         
         \multirow{2}{*}{\makecell[l]{\bf \systemname{} \\ \hspace{0.5cm} +  \bf \textsc{contra}}}  & greedy & 89.6	& \textbf{69.7} & 32.0	& 20.1 & 97.6 &	82.9 &	32.5 &	27.0  & 99.0	& 56.5	& 40.8	& 24.2 \\ 
         
         & nucleus($p=0.6$) & 89.7	 & 69.2	 & 31.9 & 	20.0  & 97.7 & 83.2 & 32.2	& 26.7   &  99.0 & 55.9 &	40.7 &	23.9  \\ 
         
         
         \hline
         \hline
         \multirow{2}{*}{\makecell[l]{\bf \systemname{} \\ \hspace{0.5cm} +  \bf \textsc{clf}  + \textsc{contra}}} &greedy & 89.3 & 69.2  &  32.9 &  \textbf{20.6} & \textbf{97.8}	& \textbf{84.0} & 33.5 &	\textbf{28.1}  & \textbf{99.0}	& \textbf{56.8}	& 41.8	& 24.9  \\ 
         
         & nucleus($p=0.6$)& 89.4 & 68.6 & 32.8 & 20.4  & 97.1 & 82.6 & 33.6 & 27.4 & 99.0	& 56.0 & 41.6	& 24.4  \\ 
         

         \hline
    \end{tabular}
    }
    \caption{Evaluation of \systemname{}~with \textsc{acc} (transfer accuracy), \textsc{fl} (fluency)  and \textsc{sim} (semantic similarity), \aggmetric{} (aggregate metric). Cooperatively reducing the contrastive or the classification loss is better than \arae{}. We report the mean of five runs for our experiments. The bolded measures are the best results}
    \label{tab:main-results}
\end{table*}

\paragraph{Datasets.} We use three datasets with single attribute changes:
$i)$~\textbf{Yelp Reviews}:  business reviews listed on Yelp, labeled as either a positive or negative sentiment. $ii)$~\textbf{IMDb Movie Reviews}: 
consists of movie reviews \cite{dai-etal-2019-style} also labelled as positive or negative. $iii)$~\textbf{Political Slant}: consists of Facebook posts from the politicians of the United States Senate and the House of Representatives \cite{prabhumoye-etal-2018-style}, labeled with either democratic/republican slant. 

We provide a summary of the dataset statistics in \Cref{tab:dataset-splits}. We include datasets of varied length and complexity. Apart from having different topics,  the \imdbdata \ dataset is more formal compared to the more colloquial \yelpdata. We fix the maximum vocabulary size for \yelpdata{}, \imdbdata{} and \politicaldata{} at 30K which is also the default maximum vocab size used in \cite{DBLP:conf/icml/ZhaoKZRL18}.

\paragraph{Constraints:}
We constrain every sentence along \textit{six} diverse dimensions that we desire to control between the two domains: 
$i)$~\textbf{Lexical}: \textit{Sentence length} -- The transferred sentence should maintain a length similar to the original sentence (binarized to long sentences with 10 or or more words or short otherwise). $ii)$~\textbf{Syntactic}:
Presence of personal pronouns (binarized to indicate the presence of a \textit{personal pronoun}); number of adjectives (categorical up to 5); number of proper nouns (categorical up to 3); syntactic tree height (categorical up to 10). 
$iii)$~\textbf{Domain specific} -- number of \textit{domain-specific attributes} \cite{li-etal-2018-delete} (categorical up to 5). 
Further, we label the sentence with a constraint-specific, catch-all label if the bounds are beyond what we mention above. 
Since the distribution of the labels may be different, we report the F1 score on our constraints.

\subsection{Model Details}
For the encoders, we use a one-layer LSTM network with 300 hidden dimensions for all the datasets. For the critics and classification loss, we use a two-layer multi-layer perceptron with 100 hidden units. 


\noindent
\paragraph{Training Hyper-parameters:}
For all our experiments we set the learning rate of the auto-encoder ($lr_{ae}$) to 1e-3 and ($lr_{disc}$) to 1e-4. The number of discriminator steps ($n_{dis}$) is set to 5. The Adam optimizer parameters $\beta_1$=$0.5$ and $\beta_2$=$0.9$, which ensures a more conservative optimization and is known to improve stability. We also add a gradient penalty to the loss function of the discriminator that stabilizes training. All the suggestions for stabilizing training are mostly obtained from \cite{DBLP:conf/iclr/ArjovskyB17}.

\paragraph{Inference Hyper-parameters:}
We used nucleus sampling with $p \in [0.6, 0.9]$. We tried different temperatures of scaling the softmax \cite{DBLP:conf/icml/GuoPSW17} - 0.4, 0.5, 0.6, 0.7 and chose the one that produced the best result on the dev set.

\subsection{Evaluation Setup}

\paragraph{Automatic Evaluation:} Our automatic evaluation considers the following three prominent criteria: $i)$~\textbf{Semantic Similarity (\textsc{sim}):} Measured between source and translated target sentences using encoders \cite{wieting-etal-2019-beyond}, instead of \textit{n-gram} metrics like \textsc{bleu} \cite{papineni-etal-2002-bleu} which have weak correlations with human judgments. $ii)$~\textbf{Transfer Accuracy (\textsc{acc}):}  The transferred sentence should belong to the target domain and a classifier is trained to distinguish between the source and the target sentence. We use \textit{fastText} classifiers \cite{joulin2017bag} for every dataset. We achieve accuracy of $97.9$ for \yelpdata,\ $96.9$ for \imdbdata \ and $97.1$ for \politicaldata. $iii)$~\textbf{Fluency (\textsc{fl}):} A transferred sentence should be grammatically correct. We fine-tune a RoBERTa-large model on the COLA \cite{warstadt2018neural} dataset to indicate whether a sentence is linguistically acceptable. Finally, we combine the three scores into an aggregate, following the criteria suggested by \citet{krishna-etal-2020-reformulating}:
\begin{equation*}
    AGG = \frac{1}{|S|}\sum_{s \in S} \textit{\textsc{ACC} (s)} \cdot \textit{\textsc{SIM} (s)} \cdot \textit{\textsc{FL} (s)} 
\end{equation*}

\paragraph{Human Evaluation:} We also perform an indicative human evaluation where we randomly sample 100 samples from each of the three datasets and hire three researchers to rate every sentence for \flmetric{}, \simmetric{} and \accmetric{} on a 3-point scale \cite{krishna-etal-2020-reformulating}. 

\begin{figure*}[t!]
  \centering
  \includegraphics[width=\linewidth]{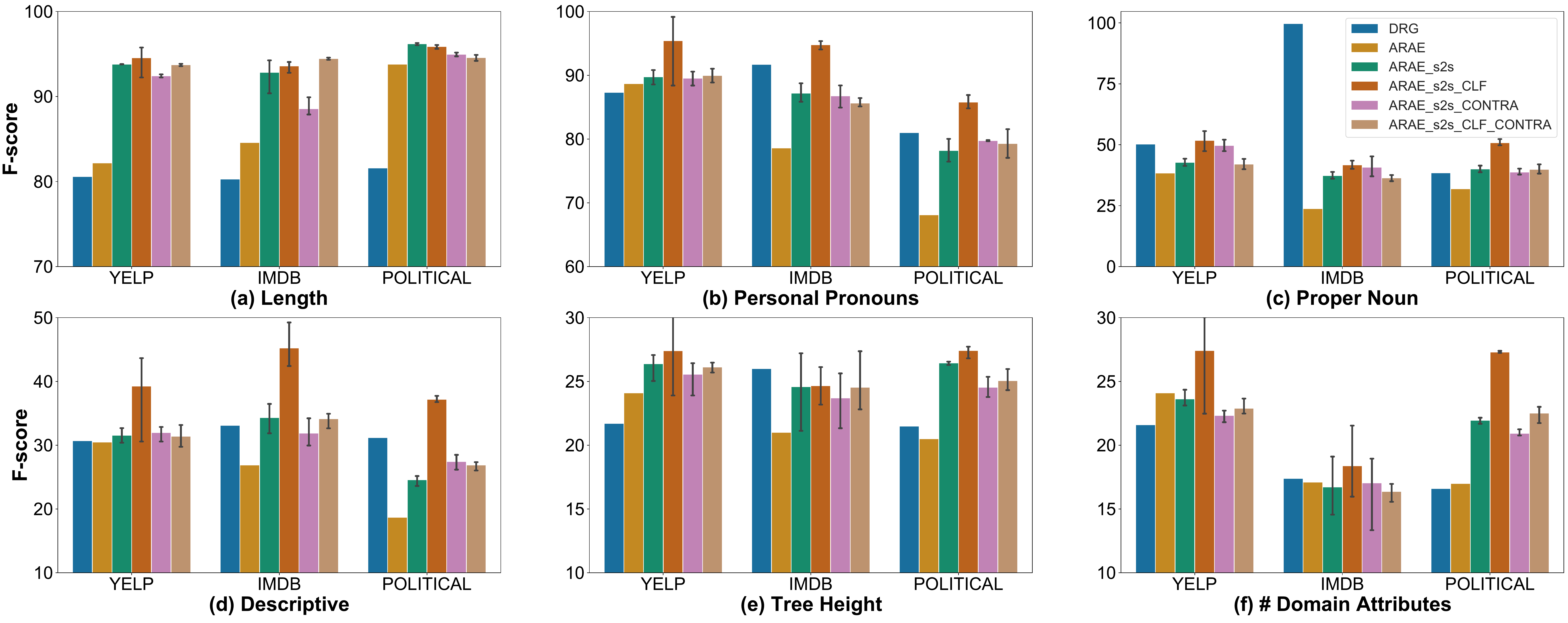}
  \caption{F-scores of different constraints. Adding cooperative losses helps in better maintaining the constraints. The error bars show the variance of generating text using greedy decoding and nucleus sampling with $p=\{0.6,0.9\}$.}
  \label{fig:maintaining-constraints}
\end{figure*}


\subsection{Baselines}
\label{sec:baselines-description}
We compare \systemname{} with the following baselines: \textbf{a) \textsc{drg}}: The Delete, Retrieve, Generate method that deletes domain specific attributes, retrieves a template and generates the target domain text \cite{li-etal-2018-delete}. We use the stronger, entire system rather than the weaker \textsc{deleteonly} and \textsc{retrieveonly} baselines; \textbf{b) \textsc{arae}}: Adversarially regularized autoencoders our system is based on \cite{DBLP:conf/icml/ZhaoKZRL18};
\textbf{c) \systemname{}}: Our model without the contrastive learning or cooperative classifier; \textbf{d) \systemname{} + \textsc{contra}}: Our model with the contrastive learning; \textbf{e) \systemname{} + \textsc{clf}}: Our  model with the cooperative classifier; \textbf{f) \systemname{}+\textsc{clf+contra}}: Our model with both the cooperative losses.
The closest model to ours is from \cite{huang-etal-2020-cycle}. However, we were not able to reproduce the results.\footnote{Repeated attempts to obtain the original source code failed.}

\subsection{Results}

\subsubsection{Overall Results}
\label{sec:overall-results}
\systemname{} + \textsc{contra} and \systemname{} + \textsc{clf} consistently perform better than \drg{} and \arae{} on the \aggmetric{} score (\Cref{tab:main-results}). The \aggmetric{} for \yelpdata{} is 20.6 (vs. 19.8), for \imdbdata{} it is 28.1 (vs. 19.9) and for \politicaldata{} 25.5 (vs. 11.0). Although cooperative loss reduction aims to satisfy the constraints between two domains, our results show that further regularization of the latent space not only brings advantages in satisfying the constraints but also improves performance~\cite{integrating-categorical-semantics-GAN}.

\textbf{Effect of Cooperative Loss Reduction on \accmetric{} \flmetric{} and \simmetric{}: } Across datasets, reducing cooperative losses improves  \accmetric{} and \flmetric{} and \simmetric{} to \arae{}. Although \drg{} produces sentences with high \simmetric{} as most of the text from the original sentence is retained after the delete step, there is a large trade-off with \accmetric{} resulting in low \aggmetric{} scores. Also, compared to \arae{}, adding cooperative losses significantly increases the \simmetric{}, with the highest increase observed for \politicaldata{}.
\abhi{The reasons for this could be two-fold: $i)$ since we mine positive sentences from a corpus that is grounded in real world events, most lexically-similar sentences may also be semantically similar}~\cite{guu-etal-2018-generating}, and $ii)$ since we tie the encoders from the source and target domain, we extract domain-agnostic information before generation, which retains content.

Fluency (\flmetric{}) also improves over all datasets. We hypothesize that reducing cooperative losses regularizes the latent space bringing fluent sentences closer together, enabling the decoder to produce semantically similar and linguistically acceptable sentences. The improvement for \politicaldata{} is less; we find these source sentences themselves are less fluent and contain many U.S. political acronyms, and that our system produces many out-of-vocabulary words affecting fluency.

\textbf{Nucleus Sampling:} Our system achieves the highest \aggmetric{} score with greedy decoding. We also experiment with nucleus sampling \cite{nucleus-sampling} with different $p$ values. We report results for only $p$=$0.6$  in \Cref{tab:main-results}, as it produced the best result. With $p$=0.6, the results are  more diverse, increasing \accmetric{} as expected. However we find that with higher values of $p$, there is a trade-off with \simmetric{} resulting in a lower \aggmetric{} score overall --- similar to \citet{krishna-etal-2020-reformulating}.

\textbf{Effect of the Number of Positives:} The number of positive and negative samples used for contrastive learning (Eq. \ref{eq:contra-loss}) have a significant effect on the overall performance \cite{DBLP:conf/nips/KhoslaTWSTIMLK20, DBLP:conf/icml/ChenK0H20, pmlr-v119-henaff20a}. \Cref{tab:ablation-study} (\textit{rows} $|P|\in \{1,2,5,10\}$) shows the \aggmetric{} scores on  \imdbdata{} (for one of the runs), for different number of positives. We find that \aggmetric{} is the highest with 2 positives per sample as also used by \citet{DBLP:conf/nips/KhoslaTWSTIMLK20}. Although increasing the number of negatives is beneficial for contrastive learning, when more than one positive example is available, using them brings further improvements \cite{DBLP:conf/nips/KhoslaTWSTIMLK20}.


\textbf{Cooperative Losses are Important on Both the Generator and Critic:}  \Cref{tab:ablation-study} shows the importance of adding the cooperative losses on the generator and critic. First, we see that adding the cooperative losses on both the generator and the critic is crucial for the overall performance. While adding the cooperative contrastive loss to both the generator and critic increases \flmetric{} and \accmetric{} while maintaining similar levels of \simmetric{}, adding the cooperative classification loss improves \simmetric{} which shows the complementary nature of the losses.  

\begin{table}[t!]
    \small 
    \centering
    \begin{tabular}{|c| c c c c|}
    \hline 
    \textbf{Model} & \textbf{\accmetric} & \textbf{\flmetric} & \textbf{\simmetric} &   \textbf{\aggmetric} \\
    \hline
    \hline
    \systemname{} + \textsc{clf} & 95.0 & 83.2 & 34.2 & 27.5 \\ 
    --~generator  & 96.2 & 87.2 & 31.3 & 26.7 \\ 
    --~critic & 94.9  & 84.4 & 30.8 & 25.5 \\
    \hline
    \hline
    \systemname{} + \textsc{contra} & 96.1 & 80.6 & 36 & 28.6 \\ 
    --~generator & 93.5 & 78.8 & 34.0 & 26.0 \\ 
    --~critic & 90.1 & 67.8  & 39.5 & 24.9 \\ 
    \hline
    \hline
    $|P|=1$ & 92.4 & 75.5 & 36.6 & 26.2  \\ 
    $|P|=2$  & 96.1 & 80.6 & 36.0 & 28.6  \\ 
    $|P|=5$  & 96.0  & 84.0  & 31.4 & 26.0  \\ 
    $|P|=10$  & 95.5 & 83.3 & 31.8 & 26.0  \\ 
    \hline 
    \end{tabular}
    \caption{Ablation study showing for cooperative losses not added to the generator (--generator) and the critic (--critic) and with different \# of positives on \imdbdata{}.}
    \label{tab:ablation-study}
\end{table}

\textbf{Human Evaluation:} We average the results and present it in \Cref{tab:human_eval}. \drg{} produces marginally better semantically similar sentences. Compared to \arae{}, our model performs well except for in \yelpdata{}. This may be because we use nucleus sampling with 0.9 which optimizes for diversity rather than similarity. On other metrics we perform on par or better than our competing systems. (See \Cref{sec:more-details-human-evaluation})

\begin{table}[t!]
    \small 
    \centering
    \begin{tabular}{|c| c| c c c|}
    \hline
    \textbf{Dataset} & \textbf{Model} & \textbf{\accmetric{}} & \textbf{\flmetric{}} & \textbf{\simmetric{}} \\
    \hline
    \multirow{3}{*}{\yelpdata{}} & \drg{}  &  2.3 &  2.1 & 2.1  \\
     &  \arae{} & \textbf{2.8} &  \textbf{2.4} &  \textbf{2.1} \\
     &  \textsc{ours} & \textbf{2.8}  & \textbf{2.4}  & 2.0  \\
    \hline
    \multirow{3}{*}{\imdbdata{}} & \drg{}  & 1.9  & 2.0  & \textbf{2.2}  \\
     &  \arae{} & 2.5  & 2.1 & 1.4 \\
     &  \textsc{ours} & \textbf{2.6}  & \textbf{2.2}  & 2.1  \\
    \hline 
    \multirow{3}{*}{\politicaldata{}} & \drg{}  & 2.3 & 2.2 & 2.1  \\
     &  \arae{} & 2.1 & 2.1 & 1.5  \\
     &  \textsc{ours} & \textbf{2.5}  & \textbf{2.4} & \textbf{2.2}  \\
    \hline
    \end{tabular}
\caption{Human evaluation of generated sentences.}
    \label{tab:human_eval}
\end{table}

\begin{table*}[]
    \centering
    \renewcommand\cellalign{lc}
    \setcellgapes{3pt}\makegapedcells
    \resizebox{\linewidth}{!}{
        \begin{tabular}{|l|l|l|l|}
        \hline
        \textbf{Dataset} & \textbf{Input} & \textbf{Output (Ours)} & \textbf{Output (\arae{})}  \\
        \hline
        \yelpdata{} & they close earlier than posted hours & they're open late night & they keep me getting better  \\
        \hline
        \imdbdata{} & \makecell[l]{this movie is a very poor attempt to \\ make money using a classical theme.} & \makecell[l]{this movie is a very good example \\ of a film that will never be forgotten.} & \makecell[l]{this is a film that has been a lot of times \\ and it's really good.} \\
        \hline
        \politicaldata{} & i wish u would bring change  & and i wish you would help bring democracy & and i 'm not sure mr.trump.  \\
        \hline
        \end{tabular}
    }
    \caption{Example outputs generated by the best system according to \aggmetric{} score.}
    \label{tab:geneartions}
\end{table*}

\textbf{Qualitative Examples:}  \Cref{tab:geneartions} shows examples of the quality of transferred examples (see \Cref{sec:more-transfer-results} for more). 
Mistakes made by the model can be attributed to poor understanding of the original semantics, lack of diversity, and 
not producing attribute-specific words.


\begin{table}[t]
    \centering
    \renewcommand\cellalign{lc}
    \setcellgapes{3pt}\makegapedcells
    \resizebox{\linewidth}{!}{
    \begin{tabular}{|l|l|l|}
    \hline 
    \textbf{Constraint} & &  \\
    \hline 
    \multirow{4}{*}{\makecell[l]{\textbf{Personal} \\ \textbf{Pronoun}}} & \makecell[l]{\textit{Source} (\imdbdata)} &\makecell[l]{jean seberg had not one iota of acting talent.} \\ 
    
    & \textit{Ours} &\makecell[l]{michael keaton was also great in his role.}\\
    
    & \textit{\arae{}} &\makecell[l]{john abraham had one of \colorbox{darkorange}{my} favorite roles .}\\
    
    
    
    
    
    \hline 
    
    \multirow{6}{*}{\makecell[l]{\textbf{Proper} \\ \textbf{Noun}}} & \makecell[l]{\textit{Source} (\imdbdata)} &\makecell[l]{\colorbox{Green1}{chris klein's} character was unlikable from \\  the start and never made an improvement} \\
    
    & \textit{Ours} & \makecell[l]{\colorbox{Green1}{robert de niro} was very good as the man \\ and she's never been}\\
    
    & \textit{\arae{}} &\makecell[l]{both of his character was made and \\ had a huge smile on me}\\

    \hline
    \end{tabular}
    }
    \caption{Table showing constraints satisfied by our system compared to \arae{}. Our method maintains constraints like number of proper nouns between sentences.}
    \label{tab:constraints-example}
\end{table}

\subsubsection{Maintaining Constraints}
\label{sec:results-maintaining-constraints}
\Cref{fig:maintaining-constraints} shows that introducing the cooperative losses significantly outperform \drg{} and \arae{} in maintaining constraints. Specifically  the \systemname{} + \textsc{clf} model performs better than \systemname + \textsc{contra}. One reason could be that, finding the appropriate positives and strong negatives can be problematic for contrastive learning. On the other hand, the classifier's objective is simpler and forces the encoder to produce representations that satisfy the different constraints effectively. 

A seemingly easy to maintain constraint is the length of the sentence. However,\ \textit{seq2seq} systems have a difficulty of maintaining appropriate lengths  \cite{murray-chiang-2018-correcting}. With no additional regularization \arae{} does not maintain the length as well as \systemname{} + \textsc{clf}. On the other hand, compared to the lexical constraints, syntactic attributes like descriptiveness, tree height and domain specific constraints present challenges, with significantly lower F scores. \systemname{} + \textsc{clf} produces significantly better results in maintaining them. This shows that obtaining improvements on the overall \aggmetric{} does not necessarily translate to producing outputs that satisfy constraints.
\drg{} maintains the proper noun for \imdbdata{} effectively, because it contains a wide variety of actor and movie names. They are retained verbatim after the delete operation. 

\begin{figure}[t]
    \centering
    \includegraphics[width=\columnwidth]{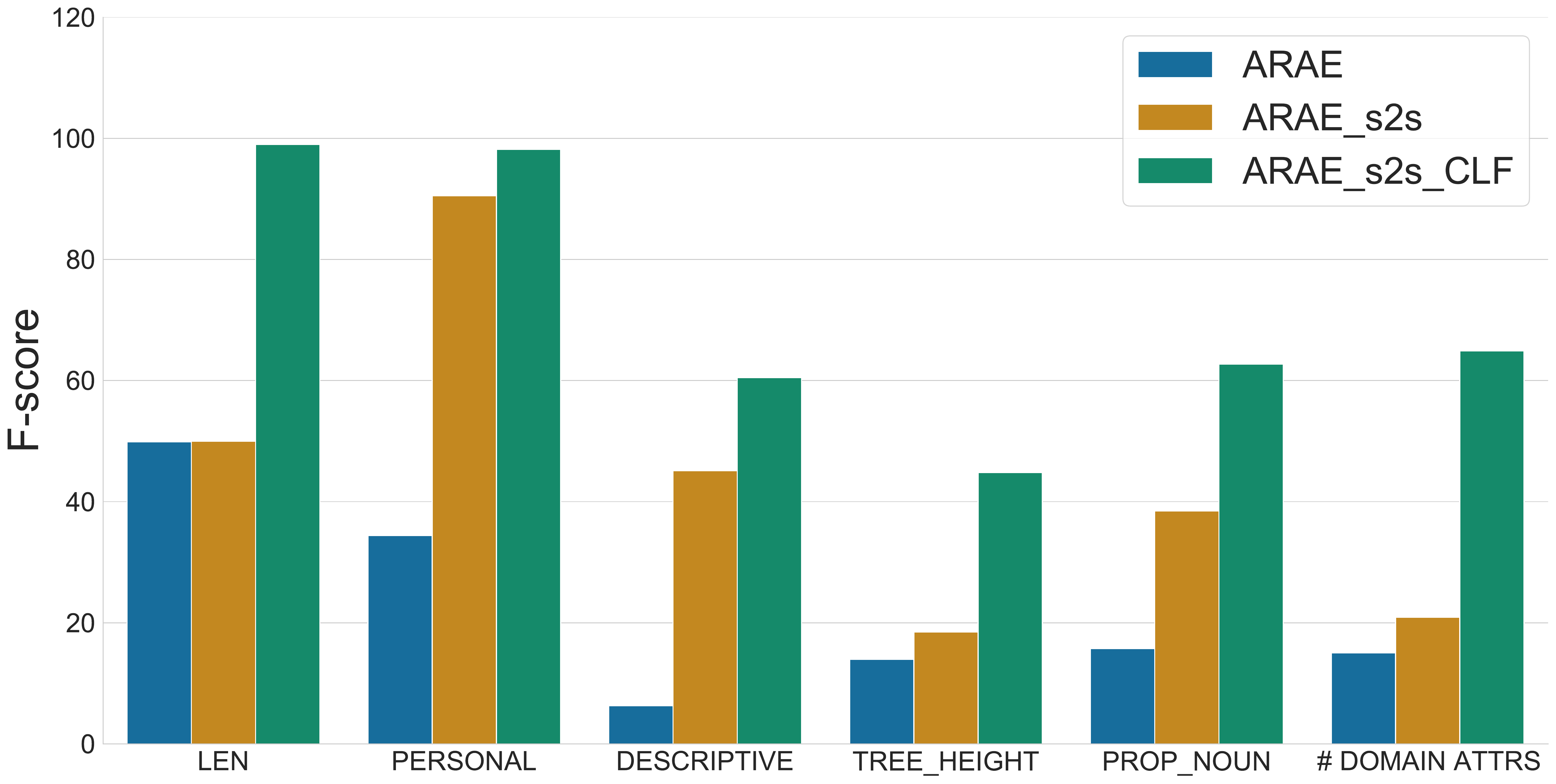}
    \caption{Comparison of \arae{}, \systemname{} and \systemname{} + \textsc{clf} for different constraints.}
    \label{fig:multi-attribute-comparison}
\end{figure}

\textbf{Multiple Attribute Datasets:}
To test whether our model can satisfy constraints across domains where multiple attributes change, we use the multi-attribute dataset released by \cite{DBLP:conf/iclr/LampleSSDRB19}. We chose the {\sc Asian} and {\sc Mexican} as two domains. Each of these domains can have multiple attributes like positive and negative sentiment text, different gender attributions to sentences, etc. We compare our \systemname{} + \textsc{clf} model with the \systemname{} and \arae{} in \Cref{fig:multi-attribute-comparison}.  The results are more pronounced in this case with \systemname{} + \textsc{clf} having clear advantage over \systemname{}. This shows that even with multiple attributes changing between domains, cooperatively reducing losses can satisfy different constraints more effectively. 

\textbf{Qualitative Examples:} 
\Cref{tab:constraints-example} shows examples of our model maintaining constraints compared to \arae{}.  Sometimes, \arae{} hallucinates and adds personal pronouns like ``my'' to the text even when there are no personal pronouns (\textit{row} 1). Also, our model produces sentences where the number of proper nouns are retained (Chris Klein vs. Robert De Niro), whereas \arae{} does not.

\section{Discussion and Limitations}
\label{section:discussion}


\paragraph{Cycle Consistency Loss:} a) \textit{In Latent Spaces} -
Cycle consistency in latent spaces has been shown to improve word-level tasks, such as cross-lingual dictionary construction \cite{mohiuddin-joty-2019-revisiting} and topic modeling \cite{hu-etal-2020-neural}. A recent work from \cite{huang-etal-2020-cycle} claims to improve unsupervised style transfer using such losses. In our experiments, however, it did not result in any noticeable performance improvement \footnote{Repeated attempts to obtain source codes failed.}. Given this, we hypothesize that cycle consistency might be too restrictive for sentence-level tasks. b) \textit{Using Back-Translation}- 
Back-translation is another alternative to ensure semantic consistency between source and the target sentence \cite{prabhumoye-etal-2018-style, artetxe2018unsupervised, lample_unsupervised_mt}. 
However, in our case, since we are training an \arae{}, it would involve an additional inference and auto-encoder training step which is expensive and we defer exploring this. 

\paragraph{Using Transformers:} We also replace our LSTM auto-encoders with both pre-trained and randomly initialized 
transformer encoder--decoders \cite{rothe-etal-2020-leveraging}. Although we found an increase in the \aggmetric{}, it was mostly because of very high \simmetric{} and very low \accmetric{}. 
Reducing the number of layers, attention heads would still result in a large model that is still prone to copying text. This reveals the potential limitations of our method and training using transformers is a  future work.




\paragraph{Transferred sentences as Adversarial Examples:}
We demonstrate an important application of our proposed constrained transfer by considering them as adversarial examples for domain adaptation. Domain Adversarial Neural Network (DANN) \cite{DBLP:series/acvpr/GaninUAGLLML17} is an unsupervised domain adaptation method that improves performance of an end-task (e.g, sentiment analysis) on a target domain considering only supervised data from source domain. We train DANN for sentiment analysis on amazon reviews dataset \cite{mcauleydataset} with {\sc DVD} as source and {\sc Electronics} as the target domain -- achieving an accuracy of 83.75\% on {\sc electronics}. 

Next, we train the best variant of \systemname{} to transfer a separate set {\sc DVD} reviews to {\sc Electronics} reviews and use them as adversarial examples to test the DANN model \footnote{Since each of {\sc DVD} and {\sc electronics} contain positive and negative reviews, we test whether transferred sentences maintain the appropriate sentiment and find the accuracy to be 79\%.}. We find that the accuracy of DANN on the {\sc electronics} domain reduces by $\sim$3 points. This shows the potential application of domain transferred sentences as adversarial examples. Similar ideas have been tried for image style transfer \cite{DBLP:journals/corr/abs-2004-12385}, but needs more investigation in NLP.


\section{Related Work}

Text attribute transfer has a vast literature \cite{dijin-survey} with deep learning methods becoming popular. The methods are either supervised (requiring parallel data) or unsupervised. Supervised methods re-purpose Sequence to Sequence models used in machine translation to achieve the goals \cite{formality-style-rao}. However, obtaining parallel data is cumbersome and thus unsupervised methods that consider pseudo-parallel data have become popular.

Disentanglement approaches are the prevalent approach to tackle unsupervised attribute transfer:  \textit{attributes} and \textit{content} are separated in latent dimension. To disentangle the attributes adversarial methods  maximize the loss of a pre-trained attribute classifier  \cite{complementray-aux-classifiers, DBLP:conf/aaai/FuTPZY18, pmlr-v80-zhao18b, john-etal-2019-disentangled}. However, the literature has paid little attention in defining and preserving content. Cycle consistency losses -- imposing that reconstruction from the target style sentence should resemble the source sentence -- is the most prevalent \cite{prabhumoye-etal-2018-style,DBLP:journals/corr/abs-1811-01135, dai-etal-2019-style, huang-etal-2020-cycle, DBLP:conf/ijcai/YiLLS20}. However, this is expensive and non-differentiable, thus requiring reinforcement learning techniques to enforce it. Our work defines the different constraints that should be preserved and adds simple differentiable contrastive learning losses to preserve them.

In recent times, text style transfer models are moving away from disentanglement approaches \cite{multiple-attribute-text-style-transfer}. Recent works that use transformers for style transfer also have adopted this \cite{dai-etal-2019-style, krishna-etal-2020-reformulating}. However, these methods do not explicitly maintain the constraints between the two styles which is the main aim of our work.

\section{Conclusion}
Text style transfer works focuses on retaining content and changing the style of sentences but does not maintain other desirable constraints. We address this by introducing two cooperative losses to the GAN-inspired Adversarially Regularized Autoencoder (ARAE) that further regularizes the latent space. While satisfying the constraints 
our methods brings significant improvements in overall score.  While we focus on simple constraints at the sentence- and word-level, future work can add phrase-level and more fine-grained constraints.
Potential future work may explore reinforcement learning losses to directly optimize the constraints.


\section*{Acknowledgments}

We would like to thank the anonymous reviewers for their useful suggestions.
We would also like to acknowledge the support of the NExT research grant funds, supported by the National Research Foundation, Prime Ministers Office, Singapore under its IRC@ SG Funding Initiative, and to gratefully acknowledge the support of NVIDIA Corporation with the donation of the GeForce GTX Titan XGPU used in this research. The work is also supported by the project no. T2MOE2008 titled CSK-NLP: Leveraging Commonsense Knowledge for NLP awarded by Singapore's Ministry of Education under its Tier-2 grant scheme.

\bibliography{custom}

\begin{thebibliography}{52}
\expandafter\ifx\csname natexlab\endcsname\relax\def\natexlab#1{#1}\fi

\bibitem[{Arjovsky and Bottou(2017)}]{DBLP:conf/iclr/ArjovskyB17}
Mart{\'{\i}}n Arjovsky and L{\'{e}}on Bottou. 2017.
\newblock \href {https://openreview.net/forum?id=Hk4\_qw5xe} {Towards
  principled methods for training generative adversarial networks}.
\newblock In \emph{5th International Conference on Learning Representations,
  {ICLR} 2017, Toulon, France, April 24-26, 2017, Conference Track
  Proceedings}. OpenReview.net.

\bibitem[{Artetxe et~al.(2018)Artetxe, Labaka, Agirre, and
  Cho}]{artetxe2018unsupervised}
Mikel Artetxe, Gorka Labaka, Eneko Agirre, and Kyunghyun Cho. 2018.
\newblock \href {http://arxiv.org/abs/1710.11041} {Unsupervised neural machine
  translation}.

\bibitem[{Artetxe et~al.(2020)Artetxe, Ruder, Yogatama, Labaka, and
  Agirre}]{artetxe-etal-2020-call}
Mikel Artetxe, Sebastian Ruder, Dani Yogatama, Gorka Labaka, and Eneko Agirre.
  2020.
\newblock \href {https://doi.org/10.18653/v1/2020.acl-main.658} {A call for
  more rigor in unsupervised cross-lingual learning}.
\newblock In \emph{Proceedings of the 58th Annual Meeting of the Association
  for Computational Linguistics}, pages 7375--7388, Online. Association for
  Computational Linguistics.

\bibitem[{Cao et~al.(2020)Cao, Shui, Pan, Kan, Liu, and
  Chua}]{cao-etal-2020-expertise}
Yixin Cao, Ruihao Shui, Liangming Pan, Min-Yen Kan, Zhiyuan Liu, and Tat-Seng
  Chua. 2020.
\newblock \href {https://doi.org/10.18653/v1/2020.acl-main.100} {Expertise
  style transfer: A new task towards better communication between experts and
  laymen}.
\newblock In \emph{Proceedings of the 58th Annual Meeting of the Association
  for Computational Linguistics}, pages 1061--1071, Online. Association for
  Computational Linguistics.

\bibitem[{Chen et~al.(2020)Chen, Kornblith, Norouzi, and
  Hinton}]{DBLP:conf/icml/ChenK0H20}
Ting Chen, Simon Kornblith, Mohammad Norouzi, and Geoffrey~E. Hinton. 2020.
\newblock \href {http://proceedings.mlr.press/v119/chen20j.html} {A simple
  framework for contrastive learning of visual representations}.
\newblock In \emph{Proceedings of the 37th International Conference on Machine
  Learning, {ICML} 2020, 13-18 July 2020, Virtual Event}, volume 119 of
  \emph{Proceedings of Machine Learning Research}, pages 1597--1607. {PMLR}.

\bibitem[{Dai et~al.(2019)Dai, Liang, Qiu, and Huang}]{dai-etal-2019-style}
Ning Dai, Jianze Liang, Xipeng Qiu, and Xuanjing Huang. 2019.
\newblock \href {https://doi.org/10.18653/v1/P19-1601} {Style transformer:
  Unpaired text style transfer without disentangled latent representation}.
\newblock In \emph{Proceedings of the 57th Annual Meeting of the Association
  for Computational Linguistics}, pages 5997--6007, Florence, Italy.
  Association for Computational Linguistics.

\bibitem[{Fu et~al.(2018)Fu, Tan, Peng, Zhao, and
  Yan}]{DBLP:conf/aaai/FuTPZY18}
Zhenxin Fu, Xiaoye Tan, Nanyun Peng, Dongyan Zhao, and Rui Yan. 2018.
\newblock \href
  {https://www.aaai.org/ocs/index.php/AAAI/AAAI18/paper/view/17015} {Style
  transfer in text: Exploration and evaluation}.
\newblock In \emph{Proceedings of the Thirty-Second {AAAI} Conference on
  Artificial Intelligence, (AAAI-18), the 30th innovative Applications of
  Artificial Intelligence (IAAI-18), and the 8th {AAAI} Symposium on
  Educational Advances in Artificial Intelligence (EAAI-18), New Orleans,
  Louisiana, USA, February 2-7, 2018}, pages 663--670. {AAAI} Press.

\bibitem[{Ganin et~al.(2017)Ganin, Ustinova, Ajakan, Germain, Larochelle,
  Laviolette, Marchand, and Lempitsky}]{DBLP:series/acvpr/GaninUAGLLML17}
Yaroslav Ganin, Evgeniya Ustinova, Hana Ajakan, Pascal Germain, Hugo
  Larochelle, Fran{\c{c}}ois Laviolette, Mario Marchand, and Victor~S.
  Lempitsky. 2017.
\newblock \href {https://doi.org/10.1007/978-3-319-58347-1\_10}
  {Domain-adversarial training of neural networks}.
\newblock In Gabriela Csurka, editor, \emph{Domain Adaptation in Computer
  Vision Applications}, Advances in Computer Vision and Pattern Recognition,
  pages 189--209. Springer.

\bibitem[{Goodfellow et~al.(2014)Goodfellow, Pouget{-}Abadie, Mirza, Xu,
  Warde{-}Farley, Ozair, Courville, and
  Bengio}]{DBLP:journals/corr/GoodfellowPMXWOCB14}
Ian~J. Goodfellow, Jean Pouget{-}Abadie, Mehdi Mirza, Bing Xu, David
  Warde{-}Farley, Sherjil Ozair, Aaron~C. Courville, and Yoshua Bengio. 2014.
\newblock \href {http://arxiv.org/abs/1406.2661} {Generative adversarial
  networks}.
\newblock \emph{CoRR}, abs/1406.2661.

\bibitem[{Guo et~al.(2017)Guo, Pleiss, Sun, and
  Weinberger}]{DBLP:conf/icml/GuoPSW17}
Chuan Guo, Geoff Pleiss, Yu~Sun, and Kilian~Q. Weinberger. 2017.
\newblock \href {http://proceedings.mlr.press/v70/guo17a.html} {On calibration
  of modern neural networks}.
\newblock In \emph{Proceedings of the 34th International Conference on Machine
  Learning, {ICML} 2017, Sydney, NSW, Australia, 6-11 August 2017}, volume~70
  of \emph{Proceedings of Machine Learning Research}, pages 1321--1330. {PMLR}.

\bibitem[{Guu et~al.(2018)Guu, Hashimoto, Oren, and
  Liang}]{guu-etal-2018-generating}
Kelvin Guu, Tatsunori~B. Hashimoto, Yonatan Oren, and Percy Liang. 2018.
\newblock \href {https://doi.org/10.1162/tacl_a_00030} {Generating sentences by
  editing prototypes}.
\newblock \emph{Transactions of the Association for Computational Linguistics},
  6:437--450.

\bibitem[{He and McAuley(2016)}]{mcauleydataset}
Ruining He and Julian~J. McAuley. 2016.
\newblock \href {https://doi.org/10.1145/2872427.2883037} {Ups and downs:
  Modeling the visual evolution of fashion trends with one-class collaborative
  filtering}.
\newblock In \emph{Proceedings of the 25th International Conference on World
  Wide Web, {WWW} 2016, Montreal, Canada, April 11 - 15, 2016}, pages 507--517.
  {ACM}.

\bibitem[{Hedayatnia et~al.(2020)Hedayatnia, Gopalakrishnan, Kim, Liu, Eric,
  and Hakkani-Tur}]{hedayatnia-etal-2020-policy}
Behnam Hedayatnia, Karthik Gopalakrishnan, Seokhwan Kim, Yang Liu, Mihail Eric,
  and Dilek Hakkani-Tur. 2020.
\newblock \href {https://www.aclweb.org/anthology/2020.inlg-1.46}
  {Policy-driven neural response generation for knowledge-grounded dialog
  systems}.
\newblock In \emph{Proceedings of the 13th International Conference on Natural
  Language Generation}, pages 412--421, Dublin, Ireland. Association for
  Computational Linguistics.

\bibitem[{Henaff(2020)}]{pmlr-v119-henaff20a}
Olivier Henaff. 2020.
\newblock \href {http://proceedings.mlr.press/v119/henaff20a.html}
  {Data-efficient image recognition with contrastive predictive coding}.
\newblock In \emph{Proceedings of the 37th International Conference on Machine
  Learning}, volume 119 of \emph{Proceedings of Machine Learning Research},
  pages 4182--4192. PMLR.

\bibitem[{Hoffer and Ailon(2015)}]{tripletloss}
Elad Hoffer and Nir Ailon. 2015.
\newblock Deep metric learning using triplet network.
\newblock In \emph{Similarity-Based Pattern Recognition}, pages 84--92, Cham.
  Springer International Publishing.

\bibitem[{Holtzman et~al.(2019)Holtzman, Buys, Forbes, and
  Choi}]{nucleus-sampling}
Ari Holtzman, Jan Buys, Maxwell Forbes, and Yejin Choi. 2019.
\newblock \href {http://arxiv.org/abs/1904.09751} {The curious case of neural
  text degeneration}.
\newblock \emph{CoRR}, abs/1904.09751.

\bibitem[{Hu et~al.(2020)Hu, Wang, Zhou, and Xiong}]{hu-etal-2020-neural}
Xuemeng Hu, Rui Wang, Deyu Zhou, and Yuxuan Xiong. 2020.
\newblock \href {https://doi.org/10.18653/v1/2020.emnlp-main.725} {Neural topic
  modeling with cycle-consistent adversarial training}.
\newblock In \emph{Proceedings of the 2020 Conference on Empirical Methods in
  Natural Language Processing (EMNLP)}, pages 9018--9030, Online. Association
  for Computational Linguistics.

\bibitem[{Huang et~al.(2020)Huang, Zhu, Xiong, Zhang, Hu, and
  Xu}]{huang-etal-2020-cycle}
Yufang Huang, Wentao Zhu, Deyi Xiong, Yiye Zhang, Changjian Hu, and Feiyu Xu.
  2020.
\newblock \href {https://doi.org/10.18653/v1/2020.coling-main.201}
  {Cycle-consistent adversarial autoencoders for unsupervised text style
  transfer}.
\newblock In \emph{Proceedings of the 28th International Conference on
  Computational Linguistics}, pages 2213--2223, Barcelona, Spain (Online).
  International Committee on Computational Linguistics.

\bibitem[{Jin et~al.(2020{\natexlab{a}})Jin, Jin, Hu, Vechtomova, and
  Mihalcea}]{dijin-survey}
Di~Jin, Zhijing Jin, Zhiting Hu, Olga Vechtomova, and Rada Mihalcea.
  2020{\natexlab{a}}.
\newblock \href {http://arxiv.org/abs/2011.00416} {Deep learning for text style
  transfer: {A} survey}.
\newblock \emph{CoRR}, abs/2011.00416.

\bibitem[{Jin et~al.(2020{\natexlab{b}})Jin, Jin, and Mihalcea}]{jin2020deep}
Di~Jin, Zhijing Jin, and Rada Mihalcea. 2020{\natexlab{b}}.
\newblock \href {http://arxiv.org/abs/2011.00416} {Deep learning for text
  attribute transfer: A survey}.

\bibitem[{John et~al.(2019)John, Mou, Bahuleyan, and
  Vechtomova}]{john-etal-2019-disentangled}
Vineet John, Lili Mou, Hareesh Bahuleyan, and Olga Vechtomova. 2019.
\newblock \href {https://doi.org/10.18653/v1/P19-1041} {Disentangled
  representation learning for non-parallel text style transfer}.
\newblock In \emph{Proceedings of the 57th Annual Meeting of the Association
  for Computational Linguistics}, pages 424--434, Florence, Italy. Association
  for Computational Linguistics.

\bibitem[{Joulin et~al.(2017)Joulin, Grave, Bojanowski, and
  Mikolov}]{joulin2017bag}
Armand Joulin, Edouard Grave, Piotr Bojanowski, and Tomas Mikolov. 2017.
\newblock Bag of tricks for efficient text classification.
\newblock In \emph{Proceedings of the 15th Conference of the European Chapter
  of the Association for Computational Linguistics: Volume 2, Short Papers},
  pages 427--431. Association for Computational Linguistics.

\bibitem[{Kang and Park(2020)}]{DBLP:conf/nips/KangP20}
Minguk Kang and Jaesik Park. 2020.
\newblock \href
  {https://proceedings.neurips.cc/paper/2020/hash/f490c742cd8318b8ee6dca10af2a163f-Abstract.html}
  {Contragan: Contrastive learning for conditional image generation}.
\newblock In \emph{Advances in Neural Information Processing Systems 33: Annual
  Conference on Neural Information Processing Systems 2020, NeurIPS 2020,
  December 6-12, 2020, virtual}.

\bibitem[{Khosla et~al.(2020)Khosla, Teterwak, Wang, Sarna, Tian, Isola,
  Maschinot, Liu, and Krishnan}]{DBLP:conf/nips/KhoslaTWSTIMLK20}
Prannay Khosla, Piotr Teterwak, Chen Wang, Aaron Sarna, Yonglong Tian, Phillip
  Isola, Aaron Maschinot, Ce~Liu, and Dilip Krishnan. 2020.
\newblock \href
  {https://proceedings.neurips.cc/paper/2020/hash/d89a66c7c80a29b1bdbab0f2a1a94af8-Abstract.html}
  {Supervised contrastive learning}.
\newblock In \emph{Advances in Neural Information Processing Systems 33: Annual
  Conference on Neural Information Processing Systems 2020, NeurIPS 2020,
  December 6-12, 2020, virtual}.

\bibitem[{Krishna et~al.(2020)Krishna, Wieting, and
  Iyyer}]{krishna-etal-2020-reformulating}
Kalpesh Krishna, John Wieting, and Mohit Iyyer. 2020.
\newblock \href {https://doi.org/10.18653/v1/2020.emnlp-main.55} {Reformulating
  unsupervised style transfer as paraphrase generation}.
\newblock In \emph{Proceedings of the 2020 Conference on Empirical Methods in
  Natural Language Processing (EMNLP)}, pages 737--762, Online. Association for
  Computational Linguistics.

\bibitem[{Lample et~al.(2017)Lample, Denoyer, and
  Ranzato}]{lample_unsupervised_mt}
Guillaume Lample, Ludovic Denoyer, and Marc'Aurelio Ranzato. 2017.
\newblock \href {http://arxiv.org/abs/1711.00043} {Unsupervised machine
  translation using monolingual corpora only}.
\newblock \emph{CoRR}, abs/1711.00043.

\bibitem[{Lample et~al.(2019)Lample, Subramanian, Smith, Denoyer, Ranzato, and
  Boureau}]{DBLP:conf/iclr/LampleSSDRB19}
Guillaume Lample, Sandeep Subramanian, Eric~Michael Smith, Ludovic Denoyer,
  Marc'Aurelio Ranzato, and Y{-}Lan Boureau. 2019.
\newblock \href {https://openreview.net/forum?id=H1g2NhC5KQ}
  {Multiple-attribute text rewriting}.
\newblock In \emph{7th International Conference on Learning Representations,
  {ICLR} 2019, New Orleans, LA, USA, May 6-9, 2019}. OpenReview.net.

\bibitem[{Lavoie{-}Marchildon et~al.(2020)Lavoie{-}Marchildon, Ahmed, and
  Courville}]{integrating-categorical-semantics-GAN}
Samuel Lavoie{-}Marchildon, Faruk Ahmed, and Aaron~C. Courville. 2020.
\newblock \href {http://arxiv.org/abs/2010.01262} {Integrating categorical
  semantics into unsupervised domain translation}.
\newblock \emph{CoRR}, abs/2010.01262.

\bibitem[{Le-Khac et~al.(2020)Le-Khac, Healy, and
  Smeaton}]{contrastive_learning_review}
Phuc~H. Le-Khac, Graham Healy, and Alan~F. Smeaton. 2020.
\newblock \href {https://doi.org/10.1109/ACCESS.2020.3031549} {Contrastive
  representation learning: A framework and review}.
\newblock \emph{IEEE Access}, 8:193907--193934.

\bibitem[{Li et~al.(2018)Li, Jia, He, and Liang}]{li-etal-2018-delete}
Juncen Li, Robin Jia, He~He, and Percy Liang. 2018.
\newblock \href {https://doi.org/10.18653/v1/N18-1169} {Delete, retrieve,
  generate: a simple approach to sentiment and style transfer}.
\newblock In \emph{Proceedings of the 2018 Conference of the North {A}merican
  Chapter of the Association for Computational Linguistics: Human Language
  Technologies, Volume 1 (Long Papers)}, pages 1865--1874, New Orleans,
  Louisiana. Association for Computational Linguistics.

\bibitem[{Li et~al.(2020)Li, Li, Zhang, Li, Zheng, Carin, and
  Gao}]{complementray-aux-classifiers}
Yuan Li, Chunyuan Li, Yizhe Zhang, Xiujun Li, Guoqing Zheng, Lawrence Carin,
  and Jianfeng Gao. 2020.
\newblock \href {https://aaai.org/ojs/index.php/AAAI/article/view/6346}
  {Complementary auxiliary classifiers for label-conditional text generation}.
\newblock In \emph{The Thirty-Fourth {AAAI} Conference on Artificial
  Intelligence, {AAAI} 2020, The Thirty-Second Innovative Applications of
  Artificial Intelligence Conference, {IAAI} 2020, The Tenth {AAAI} Symposium
  on Educational Advances in Artificial Intelligence, {EAAI} 2020, New York,
  NY, USA, February 7-12, 2020}, pages 8303--8310. {AAAI} Press.

\bibitem[{Logeswaran et~al.(2018)Logeswaran, Lee, and
  Bengio}]{DBLP:journals/corr/abs-1811-01135}
Lajanugen Logeswaran, Honglak Lee, and Samy Bengio. 2018.
\newblock \href {http://arxiv.org/abs/1811.01135} {Content preserving text
  generation with attribute controls}.
\newblock \emph{CoRR}, abs/1811.01135.

\bibitem[{Mai et~al.(2020)Mai, Pappas, Montero, Smith, and
  Henderson}]{DBLP:conf/emnlp/MaiPMSH20}
Florian Mai, Nikolaos Pappas, Ivan Montero, Noah~A. Smith, and James Henderson.
  2020.
\newblock \href {https://doi.org/10.18653/v1/2020.emnlp-main.491} {Plug and
  play autoencoders for conditional text generation}.
\newblock In \emph{Proceedings of the 2020 Conference on Empirical Methods in
  Natural Language Processing, {EMNLP} 2020, Online, November 16-20, 2020},
  pages 6076--6092. Association for Computational Linguistics.

\bibitem[{Mohiuddin and Joty(2019)}]{mohiuddin-joty-2019-revisiting}
Tasnim Mohiuddin and Shafiq Joty. 2019.
\newblock \href {https://doi.org/10.18653/v1/N19-1386} {Revisiting adversarial
  autoencoder for unsupervised word translation with cycle consistency and
  improved training}.
\newblock In \emph{Proceedings of the 2019 Conference of the North {A}merican
  Chapter of the Association for Computational Linguistics: Human Language
  Technologies, Volume 1 (Long and Short Papers)}, pages 3857--3867,
  Minneapolis, Minnesota. Association for Computational Linguistics.

\bibitem[{Murray and Chiang(2018)}]{murray-chiang-2018-correcting}
Kenton Murray and David Chiang. 2018.
\newblock \href {https://doi.org/10.18653/v1/W18-6322} {Correcting length bias
  in neural machine translation}.
\newblock In \emph{Proceedings of the Third Conference on Machine Translation:
  Research Papers}, pages 212--223, Brussels, Belgium. Association for
  Computational Linguistics.

\bibitem[{Odena et~al.(2017)Odena, Olah, and Shlens}]{acgan}
Augustus Odena, Christopher Olah, and Jonathon Shlens. 2017.
\newblock \href {http://proceedings.mlr.press/v70/odena17a.html} {Conditional
  image synthesis with auxiliary classifier gans}.
\newblock In \emph{Proceedings of the 34th International Conference on Machine
  Learning, {ICML} 2017, Sydney, NSW, Australia, 6-11 August 2017}, volume~70
  of \emph{Proceedings of Machine Learning Research}, pages 2642--2651. {PMLR}.

\bibitem[{Papineni et~al.(2002)Papineni, Roukos, Ward, and
  Zhu}]{papineni-etal-2002-bleu}
Kishore Papineni, Salim Roukos, Todd Ward, and Wei-Jing Zhu. 2002.
\newblock \href {https://doi.org/10.3115/1073083.1073135} {{B}leu: a method for
  automatic evaluation of machine translation}.
\newblock In \emph{Proceedings of the 40th Annual Meeting of the Association
  for Computational Linguistics}, pages 311--318, Philadelphia, Pennsylvania,
  USA. Association for Computational Linguistics.

\bibitem[{Prabhumoye et~al.(2018)Prabhumoye, Tsvetkov, Salakhutdinov, and
  Black}]{prabhumoye-etal-2018-style}
Shrimai Prabhumoye, Yulia Tsvetkov, Ruslan Salakhutdinov, and Alan~W Black.
  2018.
\newblock \href {https://doi.org/10.18653/v1/P18-1080} {Style transfer through
  back-translation}.
\newblock In \emph{Proceedings of the 56th Annual Meeting of the Association
  for Computational Linguistics (Volume 1: Long Papers)}, pages 866--876,
  Melbourne, Australia. Association for Computational Linguistics.

\bibitem[{Rao and Tetreault(2018)}]{formality-style-rao}
Sudha Rao and Joel~R. Tetreault. 2018.
\newblock \href {https://doi.org/10.18653/v1/n18-1012} {Dear sir or madam, may
  {I} introduce the {GYAFC} dataset: Corpus, benchmarks and metrics for
  formality style transfer}.
\newblock In \emph{Proceedings of the 2018 Conference of the North American
  Chapter of the Association for Computational Linguistics: Human Language
  Technologies, {NAACL-HLT} 2018, New Orleans, Louisiana, USA, June 1-6, 2018,
  Volume 1 (Long Papers)}, pages 129--140. Association for Computational
  Linguistics.

\bibitem[{Rothe et~al.(2020)Rothe, Narayan, and
  Severyn}]{rothe-etal-2020-leveraging}
Sascha Rothe, Shashi Narayan, and Aliaksei Severyn. 2020.
\newblock \href {https://doi.org/10.1162/tacl_a_00313} {Leveraging pre-trained
  checkpoints for sequence generation tasks}.
\newblock \emph{Transactions of the Association for Computational Linguistics},
  8:264--280.

\bibitem[{Subramanian et~al.(2018)Subramanian, Lample, Smith, Denoyer, Ranzato,
  and Boureau}]{multiple-attribute-text-style-transfer}
Sandeep Subramanian, Guillaume Lample, Eric~Michael Smith, Ludovic Denoyer,
  Marc'Aurelio Ranzato, and Y{-}Lan Boureau. 2018.
\newblock \href {http://arxiv.org/abs/1811.00552} {Multiple-attribute text
  style transfer}.
\newblock \emph{CoRR}, abs/1811.00552.

\bibitem[{Taigman et~al.(2017)Taigman, Polyak, and
  Wolf}]{DBLP:conf/iclr/TaigmanPW17}
Yaniv Taigman, Adam Polyak, and Lior Wolf. 2017.
\newblock \href {https://openreview.net/forum?id=Sk2Im59ex} {Unsupervised
  cross-domain image generation}.
\newblock In \emph{5th International Conference on Learning Representations,
  {ICLR} 2017, Toulon, France, April 24-26, 2017, Conference Track
  Proceedings}. OpenReview.net.

\bibitem[{Tian et~al.(2020)Tian, Sun, Poole, Krishnan, Schmid, and
  Isola}]{DBLP:conf/nips/Tian0PKSI20}
Yonglong Tian, Chen Sun, Ben Poole, Dilip Krishnan, Cordelia Schmid, and
  Phillip Isola. 2020.
\newblock \href
  {https://proceedings.neurips.cc/paper/2020/hash/4c2e5eaae9152079b9e95845750bb9ab-Abstract.html}
  {What makes for good views for contrastive learning?}
\newblock In \emph{Advances in Neural Information Processing Systems 33: Annual
  Conference on Neural Information Processing Systems 2020, NeurIPS 2020,
  December 6-12, 2020, virtual}.

\bibitem[{Vaswani et~al.(2017)Vaswani, Shazeer, Parmar, Uszkoreit, Jones,
  Gomez, Kaiser, and Polosukhin}]{DBLP:conf/nips/VaswaniSPUJGKP17}
Ashish Vaswani, Noam Shazeer, Niki Parmar, Jakob Uszkoreit, Llion Jones,
  Aidan~N. Gomez, Lukasz Kaiser, and Illia Polosukhin. 2017.
\newblock \href
  {https://proceedings.neurips.cc/paper/2017/hash/3f5ee243547dee91fbd053c1c4a845aa-Abstract.html}
  {Attention is all you need}.
\newblock In \emph{Advances in Neural Information Processing Systems 30: Annual
  Conference on Neural Information Processing Systems 2017, December 4-9, 2017,
  Long Beach, CA, {USA}}, pages 5998--6008.

\bibitem[{Warstadt et~al.(2018)Warstadt, Singh, and
  Bowman}]{warstadt2018neural}
Alex Warstadt, Amanpreet Singh, and Samuel~R Bowman. 2018.
\newblock Neural network acceptability judgments.
\newblock \emph{arXiv preprint arXiv:1805.12471}.

\bibitem[{Wieting et~al.(2019)Wieting, Berg-Kirkpatrick, Gimpel, and
  Neubig}]{wieting-etal-2019-beyond}
John Wieting, Taylor Berg-Kirkpatrick, Kevin Gimpel, and Graham Neubig. 2019.
\newblock \href {https://doi.org/10.18653/v1/P19-1427} {Beyond {BLEU}:training
  neural machine translation with semantic similarity}.
\newblock In \emph{Proceedings of the 57th Annual Meeting of the Association
  for Computational Linguistics}, pages 4344--4355, Florence, Italy.
  Association for Computational Linguistics.

\bibitem[{Xu et~al.(2020)Xu, Tao, Cheng, Tan, and
  Zhang}]{DBLP:journals/corr/abs-2004-12385}
Qiuling Xu, Guanhong Tao, Siyuan Cheng, Lin Tan, and Xiangyu Zhang. 2020.
\newblock \href {http://arxiv.org/abs/2004.12385} {Towards feature space
  adversarial attack}.
\newblock \emph{CoRR}, abs/2004.12385.

\bibitem[{Yi et~al.(2020)Yi, Liu, Li, and Sun}]{DBLP:conf/ijcai/YiLLS20}
Xiaoyuan Yi, Zhenghao Liu, Wenhao Li, and Maosong Sun. 2020.
\newblock \href {https://doi.org/10.24963/ijcai.2020/526} {Text style transfer
  via learning style instance supported latent space}.
\newblock In \emph{Proceedings of the Twenty-Ninth International Joint
  Conference on Artificial Intelligence, {IJCAI} 2020}, pages 3801--3807.
  ijcai.org.

\bibitem[{Yuan et~al.(2021)Yuan, Cheng, Zhang, Hao, Gan, and
  Carin}]{yuan2021improving}
Siyang Yuan, Pengyu Cheng, Ruiyi Zhang, Weituo Hao, Zhe Gan, and Lawrence
  Carin. 2021.
\newblock Improving zero-shot voice style transfer via disentangled
  representation learning.
\newblock \emph{arXiv preprint arXiv:2103.09420}.

\bibitem[{Zhao et~al.(2018{\natexlab{a}})Zhao, Kim, Zhang, Rush, and
  LeCun}]{pmlr-v80-zhao18b}
Junbo Zhao, Yoon Kim, Kelly Zhang, Alexander Rush, and Yann LeCun.
  2018{\natexlab{a}}.
\newblock \href {http://proceedings.mlr.press/v80/zhao18b.html} {Adversarially
  regularized autoencoders}.
\newblock In \emph{Proceedings of the 35th International Conference on Machine
  Learning}, volume~80 of \emph{Proceedings of Machine Learning Research},
  pages 5902--5911, Stockholmsmässan, Stockholm Sweden. PMLR.

\bibitem[{Zhao et~al.(2018{\natexlab{b}})Zhao, Kim, Zhang, Rush, and
  LeCun}]{DBLP:conf/icml/ZhaoKZRL18}
Junbo~Jake Zhao, Yoon Kim, Kelly Zhang, Alexander~M. Rush, and Yann LeCun.
  2018{\natexlab{b}}.
\newblock \href {http://proceedings.mlr.press/v80/zhao18b.html} {Adversarially
  regularized autoencoders}.
\newblock In \emph{Proceedings of the 35th International Conference on Machine
  Learning, {ICML} 2018, Stockholmsm{\"{a}}ssan, Stockholm, Sweden, July 10-15,
  2018}, volume~80 of \emph{Proceedings of Machine Learning Research}, pages
  5897--5906. {PMLR}.

\bibitem[{Zhu et~al.(2017)Zhu, Park, Isola, and
  Efros}]{DBLP:conf/iccv/ZhuPIE17}
Jun{-}Yan Zhu, Taesung Park, Phillip Isola, and Alexei~A. Efros. 2017.
\newblock \href {https://doi.org/10.1109/ICCV.2017.244} {Unpaired
  image-to-image translation using cycle-consistent adversarial networks}.
\newblock In \emph{{IEEE} International Conference on Computer Vision, {ICCV}
  2017, Venice, Italy, October 22-29, 2017}, pages 2242--2251. {IEEE} Computer
  Society.

\end{thebibliography}
\bibliographystyle{acl_natbib}

\clearpage
\appendix

\section{Transfer Results}
\label{sec:more-transfer-results}
More transfer results are mention in \Cref{tab:more_positive_examples}. Examples where our system fails with plausible explanation are given in \Cref{tab:negative_examples}. Examples of translation from the multi-attribute dataset is shown in \Cref{tab:compositional_data_examples}.

\section{More details on Human Evaluation}
\label{sec:more-details-human-evaluation}
For \flmetric{}, 0 indicates not fluent at all, 1 indicates somewhat fluent and 2 is a completely fluent sentence. We explicitly ask the annotators to consider semantic similarity for \simmetric{}, irrespective of whether the target sentence shares some phrases with the source sentence, with 1 indicating no semantic similarity and 3 indicating complete semantic similarity. For \accmetric{}, 1 indicates that the target sentence has only the source sentence style while 2 indicates good transfer to the target style. 

\begin{table}[h!]
    \small 
    \centering
    \begin{tabular}{|c c c|}
    \hline 
    \textbf{Dataset} & \textbf{Metric} & $\alpha$ \\
    \hline
    \multirow{3}{*}{\yelpdata{}} & \accmetric{} &  0.69   \\
         & \flmetric{} & 0.33 \\
         & \simmetric{} & 0.49  \\
    \hline
    \multirow{3}{*}{\imdbdata{}} & \accmetric{} &  0.60  \\
         & \flmetric{} &  0.38  \\
         & \simmetric{} & 0.48  \\
    \hline 
    \multirow{3}{*}{\politicaldata{}} & \accmetric{} & 0.76   \\
         & \flmetric{} & 0.71  \\
         & \simmetric{} & 0.71  \\
    \hline 
    \end{tabular}
    \caption{Krippendorff's alpha showing inter annotator agreement for \yelpdata{}, \imdbdata{} and \politicaldata{}}
    \label{tab:krippendorffs_alpha}
\end{table} 

We calculate the Krippendorff's alpha to assess the inter annotator agreement.  \Cref{tab:krippendorffs_alpha} shows the inter-annotator agreement. An $\alpha$ of 0.4 is considered good agreeement \cite{hedayatnia-etal-2020-policy}. We have moderate to good agreements on all the datasets for different measures. On more inspection we found that the disagreements in fluency mostly arrives for small phrases like "my fav" although is an accepted phrase in social media text is considered 2 by one annotator and 3 by another. We also further note that, smaller sentences were easier to judge and had better agreement rates on \simmetric{} compared to longer sentences. 

\textbf{Information about participants:} We hire three graduate researchers in NLP (average age ~25) for the annotation task who are well versed in English. We obtained permission for their participation and compensated them appropriately according to hourly wages in the country. The specific instruction given to them for the evaluation are as follows.

\noindent 
Consider two sentences 

\begin{itemize}
    \item \textbf{Source sentence}: Sentence from the source domain
    \item \textbf{Target sentence}: The transferred sentence produced by one of the systems
\end{itemize}

For every target sentence you will be asked to rate it according to three measures described below. 
\smallskip 

\noindent 
\textbf{Fluency}: Indicate how fluent the target sentence is (regardless of whether the sentence is appropriately transferred to the target sentence)
\smallskip

1 -  Not fluent at all - Does not look like an English sentence.

2 -  Fluent but with some mistakes - Fluent but with some grammatical errors 

3 -  Entirely fluent. - A good English Sentence 

\smallskip

\noindent 
\textbf{Similarity}: Indicate how semantically similar the target sentence is.
\smallskip

\noindent 
1 - Does not share any words/phrases with the source sentence and/or is not semantically similar (does not share high level topics of the sentence)

\noindent 
2 - Shares some words/phrases with the source sentence and/or has moderate level of semantic similarity (talks about similar high level topics)

\noindent 
3 - Shares appropriate  words/phrases with the source sentence and is highly semantically similar

\noindent 
\textbf{Accuracy}: Indicate whether the target sentence is accurately transferred to the target domain

\textbf{Sentiment Transfer}

1 - The target sentiment is not evident in the target sentence at all. Has words expressing opposite sentiment 

2 - Neutral Sentiment. Choose this option, if it has both positive and negative sentiment

3 - The target sentiment is evident in the target sentiment. Has appropriate sentiment bearing words.

If the sentence itself has no sentiment then chose 2

\smallskip 
\noindent 
\textbf{Political Orientation}
\smallskip 

1 - Talks about topics with the other orientation. For example, if the target style is democratic and the target sentence talks about conservative issues like abortion, gun control

2 - Neutral. 

3 - Talks about topics with the correct orientation. For example, if the target style is democratic and talks about progressive issues like liberty, free speech, Elizabeth Warren, Joe Biden, gay rights etc.

\begin{table*}[t]
    \adjustbox{max width=\textwidth}{
    \begin{tabular}{|c p{7cm} p{7cm}|}
    \hline
    \textbf{Dataset} & \textbf{Source} & \textbf{Target}  \\
    \hline
    \yelpdata{} & consistently slow. & consistently good. \\
    \yelpdata{} & so nasty. & so delicious! \\
    \yelpdata{} & i hate mayonnaise. & i love chipotle! \\
    \yelpdata{} & i 'm so disappointed! & i 'm so impressed! \\
    \yelpdata{} & but service was horrible both times. & but service was really good \& fast. \\
    \yelpdata{} & now the service i experienced was bad. & now i have the best service.\\
    \yelpdata{}& the chicken tenders did n't taste like chicken  &  wtf?,the chicken marsala , really good tomato , love! \\
    \yelpdata{}& the food was nothing special and the service was slow. & the food was amazing , the service is good. \\
    \yelpdata{}& that's why i think its shady . & that's why i think its finest. \\    
    \yelpdata{}& that stuff was awful. & that's delicious! \\
    \yelpdata{}& disgusting all around. & great , all around. \\
    \yelpdata{}& the rice was dry. & the rice was delicious. \\
    \yelpdata{}& the sweet and sour chicken is hit and miss. & the sweet and sour chicken is a winner here. \\
    \hline
    \imdbdata{}& the dialog is poorly written  &  the writing and direction are so precise, and he captures the spirit. \\
    \imdbdata{}& i'm a sucker for a good pirate movie, but this ain't it. & i'm a huge fan of the genre , but this movie is definitely worth it. \\
    \imdbdata{}& don't see this movie. & don't miss this movie. \\
    \imdbdata{}& terrible movie made on zero budget. & absolutely amazing movie on tv. \\
    \imdbdata{}& maybe the worse movie i have ever see. & maybe the best movie i have ever seen. \\
    \imdbdata{}& never would i recommend this movie to my worst enemy, yet anybody i actually like. & i would recommend this movie to anyone who enjoys good wholesome, clean fun. \\
    \imdbdata{}& tedious, not hilarious. & real, great. \\
    \imdbdata{}& this movie is truly one of the worst movies i 've ever seen. & this movie is one of the best movies i 've ever seen. \\
    \imdbdata{}& it was one of the shortest movies i 've ever seen, and thank god! & it was one of the most original films i've ever seen, and i'm glad. \\
    \imdbdata{}& do not watch this movie sober. & do not miss this movie. \\
    \imdbdata{}& wesley snipes is a far more accomplished actor than to be in this. & rob roy is a great actor in his own right to date. \\
    \imdbdata{}& this film is a real yawner. & this film is a true delight. \\
    \imdbdata{}& my rating : 2/10. & my vote : 9/10. \\
    \imdbdata{}& some competent acting talent was squandered. & an excellent performance by everyone.  \\
    \hline
    \politicaldata{}& support you, rand. & support you, elizabeth. \\
    \politicaldata{}& borders first. & equal rights \\
    \politicaldata{}& keep telling yourself that  &  ted.,keep telling that truth, keith. \\
    \politicaldata{}& just love the constitution. & just love the dnc. \\
    \politicaldata{}& for supporting clemson and for working for a balance budget . & for supporting student loans  for a working and fair job. \\
    \politicaldata{}& for you service trey ! & for you service kamala! \\
    \politicaldata{} & save america! & save us elizabeth \\
    \politicaldata{} & stand with your constituents and vote to defund obama care.  & stand with your constituents and vote for bernie sanders' bill!! \\
    \politicaldata{} & poliquin has been a strong voice for the people of northern maine he has my vote & carol has been doing a great job for the people of this state of ohio. \\
    \bottomrule[0.5pt]
    \end{tabular}
    }
    \caption{More examples from our best performing model for \yelpdata{}, \imdbdata{}, \politicaldata{}}
    \label{tab:more_positive_examples}
\end{table*}

\begin{table*}[t]
    \adjustbox{max width=\textwidth}{
    \centering 
    \begin{tabular}{|c p{3cm} p{3cm} p{5cm}|}
    \hline 
    \textbf{\textbf{Dataset}} & \textbf{\textbf{Source}} & \textbf{\textbf{Target}}  & \textbf{Explanation} \\
    \hline
    \yelpdata & completely outdated, old hotel. & completely charming and old school. & The model produces mixed sentiments without understanding that ``old school'' has negative connotations \\ 
    \yelpdata{} & bad service, bad food. & great food, amazing food. & Lack of diversity in the generation and the model does not produce outputs with respect to service \\
    \hline 
    \imdbdata{} & music is boring, and starts to annoy after 15-20 minutes. & its an epic and very moving film, without being preachy. & The model fails to produce semantically similar sentence. Probably because music is not a frequent topic in the dataset \\ 
    \imdbdata{} & brad pitt overacts appallingly. & john woo does it. & Although the the model reproduces a name, it does not produce a fluent sentence \\ 
    \hline 
    \politicaldata{} & obamacare, no one wants it!! & al, no one cares it. & Does not understand that "Obamacare" is an entity and hallucinates and uses "care" as a verb \\ 
    \politicaldata{} & are clearly not republican anymore! & are not enough sen booker. & Hallucinates Sen Booker which appears frequently in the dataset \\
    \hline
    \end{tabular}
    }
    \caption{Mistakes that \systemname{} makes and plausible explanations}
    \label{tab:negative_examples}
\end{table*}

\begin{table*}[]
    \centering
    \adjustbox{max width=\textwidth}{
    \begin{tabular}{|p{8cm} p{8cm}|} 
    \hline 
    \textbf{Source} & \textbf{Target} \\
    \hline 
     by far, the best spot for ramen. simple menu. fast service. silky, creamy chicken broth. & by far the best breakfast tacos in the area. friendly staff. great food. ask for the spicy chicken, and they have a great selection.  \\
     \hline 
     
     try sushi boat. it's totally amazing. they offer good food and high quality. good sake is ready. thank you for good place. & love it. good food. they have good margaritas and good food. good prices. there's a good amount of food for you. \\ 
     \hline 
     
     best thai in austin. we love the atmosphere, the service and obviously the food. they did a great job last time we were there since our party had specific requirements like <unk> free and <unk>. & best mexican food in the area. the service was great and the food was so good. we had a party of 10 and they were very accommodating to our group of us. we were there last night and the food was good \\ 
     \hline 
     
     fabulous, delicious, authentic. at lunch on a saturday the place was packed! 20 minute wait for a table. i was one of two customers who was not chinese. i 'll be back frequently. & awesome mexican food, a little on the corner of a <unk>. i was here on a saturday night. they were busy, but we were able to get a table. i will definitely be back! \\ 
     \hline 
     
     this place is great! i grew up going to china inn in chamblee plaza and it's the same owner! lunch service is fast and delicious! give it a shot, you won't be disappointed ! & this place is awesome!! i've been coming to this location for years and it's always clean and the service is fast and friendly. it's a great mexican restaurant, you can't go wrong with the food! \\ 
     \hline 
     
     awful. i'm writing this as i eat it now. worst poke bowl i've ever had. the smallest portion of poke possible, <unk> overcooked rice, and barely got any ponzu. most standard toppings cost extra too.  & awful! i've never had a bad meal here. i only ordered two of them. the only thing i didn't like was the <unk>. it's not much flavor, but the meat is dry. \\
     \hline 
     
     worst chinese food experience i ever had. told the manager about my allergies and that all i wanted was vegetable fried rice no soy sauce they couldn't even handle that!!! amateur hour here don't waste your time. go to china blossom & worst experience ever. i ordered the <unk> and they were all wrong with that i couldn't eat the food. that's how i don't care about how they charge you for the fajitas. no one ever came to eat here. \\ 
     \hline 
     
     the food was terrible. it definitely was not fresh. the broccoli was over cooked on my beef broccoli. my chicken chow mean fried rice just looked and tasted like last weeks rice. there was one chunk of chicken and <unk> pieces of egg in & the food was just ok. the chicken was dry. it was very dry. i ordered the chicken chimichanga and it was just plain gross. the only thing that was <unk> was the chicken burrito. there was only one other person in the <unk> \\ 
     \hline 
     
    \end{tabular}
    }
    \caption{Examples for multiple-attribute dataset} 
    \label{tab:compositional_data_examples}
\end{table*}

\end{document}